\documentclass[
	article,			% indica que é um artigo acadêmico
	11pt,				% tamanho da fonte
	oneside,			% para impressão apenas no verso. Oposto a twoside
	a4paper,			% tamanho do papel.
	english,			% idioma adicional para hifenização
	sumario=tradicional
	]{abntex2}

\usepackage{lmodern}			% Usa a fonte Latin Modern
\usepackage[T1]{fontenc}		% Selecao de codigos de fonte.
\usepackage[utf8]{inputenc}		% Codificacao do documento (conversão automática dos acentos)
\usepackage{indentfirst}		% Indenta o primeiro parágrafo de cada seção.
\usepackage{nomencl} 			% Lista de simbolos
\usepackage{color}				% Controle das cores
\usepackage{graphicx}			% Inclusão de gráficos
\usepackage{microtype} 			% para melhorias de justificação
\usepackage{amsmath,amsthm,amsfonts,amstext,amscd,bezier,amssymb}

\usepackage{ae,aecompl}
\usepackage{url}
\usepackage{graphicx}
\usepackage{subfig}
\usepackage{multirow}
\usepackage{epstopdf}
\usepackage{empheq}
\usepackage{setspace}
\usepackage{booktabs}
\usepackage{textcomp}
\usepackage{pdfpages}
\usepackage[vlined, boxed, english, algoruled, longend, commentsnumbered]{algorithm2e}
\usepackage{listings}
\usepackage{latexsym}
\usepackage[mathcal]{eucal}
\usepackage{blindtext}
\usepackage{colortbl}
\usepackage{placeins}
\usepackage{caption}
\usepackage{tabularx}
\usepackage{svg}
\usepackage{amsthm}
\usepackage{gensymb} % \degree
\usepackage[
			hyperpageref]{backref}	 % Paginas com as citações na bibl
\usepackage[alf]{abntex2cite}	% Citações padrão ABNT
\newtheorem{theorem}{Theorem}
\newcommand{\M}{\text{$\mathcal M$}}
\newcommand{\R}{\text{$\mathbb R$}}

\renewcommand*{\backrefalt}[4]{
	\ifcase #1 %
		No citing on text.%
	\or
		Cited on page #2.%
	\else
		Cited #1 times on pages #2.%
	\fi}%
\titulo{Applying Lie Groups Approaches for Rigid Registration of Point Clouds}
\autor{Liliane Rodrigues de Almeida (1), Gilson A. Giraldi(1),\\ Marcelo Bernardes Vieira(2)\\ \\
	   (1)Laboratório Nacional de Computação Científica \\
        Petrópolis, RJ, Brasil.\\
	   lilianera@lncc.br , gilson@lncc.br \\
	   (2) Departamento de Ci\^{e}ncia da Computa\c{c}\~{a}o\\
	   Universidade Federal de Juiz de Fora\\
       marcelo.bernardes@ufjf.edu.br}
\date{December 12, 2018}
\definecolor{blue}{RGB}{41,5,195}

\makeatletter
\hypersetup{
		pdftitle={\@title},
		pdfauthor={\@author},
    	pdfsubject={\imprimirpreambulo},
	    pdfcreator={LaTeX with abnTeX2},
		pdfkeywords={abnt}{latex}{abntex}{abntex2}{trabalho acadêmico},
		colorlinks=true,       		% false: boxed links; true: colored links
    	linkcolor=blue,          	% color of internal links
    	citecolor=blue,        		% color of links to bibliography
    	filecolor=magenta,      		% color of file links
		urlcolor=blue,
		bookmarksdepth=4
}
\makeatother
\makeindex
\setlrmarginsandblock{3cm}{3cm}{*}
\setulmarginsandblock{3cm}{3cm}{*}
\checkandfixthelayout
\SingleSpacing

\begin{document}

% Seleciona o idioma do documento (conforme pacotes do babel)
\selectlanguage{english} % se não selecionar aqui, a data fica em portugues.
%\selectlanguage{brazil}

% Retira espaço extra obsoleto entre as frases.
\frenchspacing

% ----------------------------------------------------------
% ELEMENTOS PRÉ-TEXTUAIS
% ----------------------------------------------------------

%---
%
% Se desejar escrever o artigo em duas colunas, descomente a linha abaixo
% e a linha com o texto ``FIM DE ARTIGO EM DUAS COLUNAS''.
% \twocolumn[    		% INICIO DE ARTIGO EM DUAS COLUNAS
%
%---
% página de titulo
\maketitle

% resumo em inglês
\begin{resumo}[Abstract]
 \begin{otherlanguage*}{english}
In the last decades, some literature appeared using the
Lie groups theory to solve problems in computer vision. On the other hand, Lie algebraic representations of the transformations therein were introduced to overcome the difficulties behind group structure by mapping the transformation groups to linear spaces. In this paper we focus on application of Lie groups and Lie algebras to find the rigid transformation that best register two surfaces represented by point clouds. The so called pairwise rigid registration can be formulated by comparing intrinsic second-order orientation tensors that encode local geometry. These tensors can be (locally) represented by symmetric non-negative definite matrices. In this paper we interpret the obtained tensor field as a multivariate normal model. So, we start with the fact that the space of Gaussians can be equipped with a Lie group structure, that is isomorphic to a subgroup of the upper triangular matrices. Consequently, the associated Lie algebra structure enables us to handle Gaussians, and consequently, to compare orientation tensors, with Euclidean operations. We apply this methodology to variants of the Iterative Closest Point (ICP), a known technique for pairwise registration.
We compare the obtained results with the original implementations that apply the comparative tensor shape factor (CTSF), which is  a similarity notion based on the eigenvalues of the orientation tensors. We notice that the similarity measure in tensor spaces directly derived from Lie's approach is not invariant under rotations, which is a problem in terms of rigid registration. Despite of this, the performed computational experiments show promising results when embedding orientation tensor fields in Lie algebras.

%The focused methods  apply the comparative tensor shape factor (CTSF) in their implementations, which is  a similarity %notion based on the eigenvalues of the orientation tensors, generating the ICP-CTSF and Shape-based Weighting %Covariance ICP (SWC-ICP). We replace the CTSF
   \vspace{\onelineskip}

 \end{otherlanguage*}
\end{resumo}

% ---

% ----------------------------------------------------------
% ELEMENTOS TEXTUAIS
% ----------------------------------------------------------
\textual

\section{Introduction}
In computer vision application like robotics, augmented reality, photogrammetry, and pose estimation, the dynamical analysis can be formulated using
transformation matrices as well as some kind of matrix operation \cite{Zhang:2014}. These ingredients comprise
the elements of algebraic structures
known as Lie groups, which are differentiable manifolds provided with an operation that satisfies group axioms and whose tangent space at the
identity defines a Lie algebra \cite{Huang2017DeepLO}. One of the advantages of going from Lie group to Lie algebra is that we can replace the multiplicative structure to an equivalent
vector space representation, which makes statistical learning methods and correlation calculation more rational and precise \cite{Xu2012ApplicationsOL}.

Surface registration is a common problem in the mentioned applications  \cite{berger:2016}, encompassing rigid registration that deals only with sets that differ by a rigid motion. In this case, given two point clouds, named source set $P = \{\mathbf{p}_i | \mathbf{p}_i=(p_{ix},p_{iy},p_{iz})\}$, and target set $Q = \{ \mathbf{q}_j | \mathbf{q}_j=(q_{jx},q_{jy},q_{jz})\}$, we need to find a motion transformation $T$, composed by a rotation $R$ and a translation $\mathbf{t}$, that applied to $P$ best align both clouds ($T(P) \approx Q$), according to a distance metric.

The classical and most cited algorithm in the literature to rigid registration is the Iterative Closest Point (ICP) \cite{Besl}. This algorithm takes as input the point clouds $P$ and $Q$, and consists of the iteration of two major steps: matching between the point clouds and transformation estimation. The matching searches the closest point in $P$ for every point in $Q$ . This set of correspondences is used to estimate a rigid transformation. These two steps are iterated until a termination criterion is satisfied.

In order to compute the correspondence set, ICP and many other registration techniques use just the criterion of minimizing point-to-point Euclidean distances between the sets $P$ and $Q$. This approach might not be efficient in cases of partial overlapping, because only a subset of each point cloud has a correct correspondent instead of all the points.
Local geometric features can be used to enhance the quality of the matching step. The key idea is to compute descriptors which efficiently encode the geometry of a region of the point cloud \cite{Sharp}. Considering that we do not have triangular meshes representing the surfaces, we choose intrinsic elements that can be computed using the neighborhood of a point $\mathbf{p}$, that is represented by the list $L_{k}(\mathbf{p})$ of its $k$ nearest-neighbors. Specifically, Cejnog \cite{Cejnog:2017} proposed a combination between the ICP and the comparative tensor shape factor (ICP-CTSF) method that is based on second-order orientation tensors, computed using $L_{k}(\mathbf{p})$, which encodes the local geometry nearby the point $\mathbf{p}$. As in the focused application we have different viewpoints, with close perspectives of the same object, we expect that two correspondent points belonging to different viewpoints would have similar neighborhood disposition. Consequently, they are likely to have associated tensors with similar eigenvalues. This is the fundamental assumption in Cejnog's work that has shown greatly enhancement in the matching step when combined with the closest point searching \cite{Cejnog:2017}.

%Behind this assumption we have the idea of comparing orientation tensors and the source and target cloud positions.

More precisely, the above process generates two (discrete) tensor fields, say $\mathbf{S}_{1}:P\rightarrow \mathbb{R}^{3\times3}$ and $\mathbf{S}_{2}:Q\rightarrow\mathbb{R}^{3\times3}$. Given two points $\mathbf{p} \in P$ and $\mathbf{q} \in Q$ the Cejnog's assumption is based on the idea of computing a similarity measure between two symmetric tensors  $\mathbf{S}_{1}(\mathbf{p})$ and $\mathbf{S}_{2}(\mathbf{q})$ by comparing their eigenvalues. Then, the similarity value is added to the Euclidean distance between $\mathbf{p}$ and $\mathbf{q}$ to get the correlation between them. Such viewpoint forwards some theoretical issues regarding tensor representation and metrics in the space of orientation tensors.

Given a smooth manifold $\M$ of dimension $n$, a tensor field can be formulated through the notion of tensor bundle which in turn is the set of pairs $(\mathbf{x},\mathbf{S})$, where  $\mathbf{x} \in \M$ and $\mathbf{S}$ is any tensor defined over $\M$. In terms of parametrization, it is straightforward to show that the tensor bundle can be covered by open sets that are isomorphic to open sets of $\mathbb{R}^{m}$ where $m$ depends on $n$ and the tensor space dimension in each point $\mathbf{x} \in \M$ \cite{Renteln:2013}.  In this sense, the tangent bundle is a differentiable manifold of dimension $m$ itself. Moreover, we can turn it into a Riemannian manifold by using a suitable metric which, for the purpose of statistical learning, can be the Fisher information one in the case of orientation tensors \cite{stanford1981riemannian}. Therefore, we get the main assumption of this paper:  compute matching step using the Riemannian structure behind the tensor bundle.

However, once orientation tensors can be represented by symmetric, non-negative definite matrices, we can embed our assumption in a multivariate Gaussian model and apply the methodology presented in \cite{Li2017LocalLM}, which shows that the space of Gaussians can be equipped with a Lie group structure. The advantage of this viewpoint is that we can apply the corresponding Lie algebraic machinery  which enables us to handle Gaussians (in our case, orientation tensors) with Euclidean operations
rather than complicated Riemannian operations. More specifically, we show that the Frobenius norm in the Lie algebra of the orientation tensors over $\M$ combines  both the position of the points and the local geometry.

In this study, we focus on the ICP-CTSF and the combination between the Shape-based Weighting Covariance ICP and the CTSF, named SWC-ICP \cite{GilsonAkio2018}. We keep the core of each algorithm and insert in the matching step the Frobenius norm in the Lie space.
In the tests, we analyze the focused techniques against fundamental issues in registration methods \cite{Salvi, Tam, Diez}. Firstly, some of them, like ICP, assume that there is a correct correspondence between the points of both clouds. This assumption easily fails on real applications because, in general, the acquired data is noisy and we need to scan the object from multiple directions, due to self-occlusion as well as limited sensor range, producing only partially overlapped point clouds. Another issue of ICP and some variants is that they expect that the point clouds are already coarsely aligned. Besides, missing data is a frequent problem, particularly when using depth data, due uncertainty caused by reflections in the acquisition process.

To evaluate each algorithm in the target application, we consider point clouds acquired through a Cyberware 3030 MS scanner \cite{CyberwareScanner} available in the Stanford $3D$ scanning repository \cite{Stanford}. The Bunny model  was chosen for the tests. We take the original data and generate multiple case-studies, through a rigid transformation, addition of noise and missing data. In this case, the ground truth rotation and translation are available and, as a consequence, we can  measure the errors without ambiguities.

The remainder of this paper is organized as follows. Section \ref{sec:Registration-Algs}, we
summarize the considered registration methods as well as the traditional ICP to set the background for rigid registration. Next, section \ref{sec:Lie} describes basic concepts in Lie groups and Lie algebras. The technique for embedding Gaussian models in linear spaces (named DE-LogE) is revised in \ref{sec:Embed}. The corresponding logarithm expression is analyzed in more details in section \ref{sec:DE-LogE}. Next, section \ref{sec:Comparing-Tensors} discuss the comparison between orientation tensors in the associated Lie algebra while section \ref{sec:effect-rigidT} analyses the influence of rigid transformations.
The section \ref{sec:Experimental-Results} shows the experimental results
obtained by applying DE-LogE in the registration methods to point clouds.
Section \ref{sec:concl} presents the conclusions and future researches.

\section{Registration Algorithms \label{sec:Registration-Algs}}

We apply in this work two different algorithms to rigid registration:  ICP-CTSF \cite{Cejnog:2017} and SWC-ICP \cite{Yamada}
In this section, we aim to establish the necessary notation and the
mathematical formulation behind these techniques.

Hence, the bold uppercase symbols represent tensor objects,
such as $\mathbf{T},\mathbf{S}$; the normal uppercase symbols represent
matrices, data sets and subspaces ($P$, $U$, $D$, $\Sigma$, etc.); the bold
lowercase symbols denotes vectors (represented by column arrays) such
as $\mathbf{x}$, $\mathbf{y}$. The normal lowercase symbols
are used to represent functions as well as scalar numbers ($f$, $\lambda$,
$\alpha$, etc.). Also, given a matrix $A\in\mathbb{R}^{m\times m}$
and a set $S$, then $tr\left(A\right)=A_{11}+A_{22}+\ldots+A_{mm}$
is the trace of $A$, and $|S|$ means the number of elements of $S$.
Besides, $I_{m}$ represents the $m\times m$ identity matrix.

Our focus is rigid registration problem in point clouds.
So, let the source and target point clouds in $\mathbb{R}^{m}$ represented,
respectively, by $P=\left\{ \mathbf{p}_{1},\mathbf{p}_{2},\ldots,\mathbf{p}_{n_{P}}\right\} \subset\mathbb{R}^{m}$ and
$Q=\left\{ \mathbf{q}_{1},\mathbf{q}_{2},\ldots,\mathbf{q}_{n_{Q}}\right\} \subset\mathbb{R}^{m}$.
A rigid transformation  $\rho:\mathbb{R}^{m}\rightarrow\mathbb{R}^{m}$
is given by:
\begin{equation}
\rho \left(\mathbf{x}\right)=R\mathbf{x}+\mathbf{t},\label{eq:rigid-transf}
\end{equation}
with $R\in SO(m)$ and $\mathbf{t}\in\mathbb{R}^{m}$ being the rotation
matrix and translation vector, respectively.

The registration problem aims at finding a rigid transformation $\rho:\mathbb{R}^{m}\rightarrow\mathbb{R}^{m}$
that brings set $P$ as close as possible to set $Q$ in terms of
a designated set distance, computed using a suitable metric $d:\mathbb{R}^{m}\times\mathbb{R}^{m}\rightarrow\mathbb{R}^{+}$,
usually the Euclidean one denoted by $d\left(\mathbf{p},\mathbf{q}\right)=\left\Vert \mathbf{p}-\mathbf{q}\right\Vert _{2}$.
To solve this task, the first step is to compute the matching relation
$C\left(P,Q\right)\subset P\times Q$ that denotes the set of all
correspondence pairs to be used as input in the procedure to compute the
transformation $\rho$. Formally, we consider:
\begin{equation}
C\left(P,Q\right)=\left\{ \left(\mathbf{x}_{i_{l}},\mathbf{y}_{i_{l}}\right)\in P\times Q;\ \mathbf{x}_{i_{l}}=\arg\min_{\mathbf{x}\in P}(d(\mathbf{x},\mathbf{y}_{i_{l}}))\right\}, \label{eq:matching00}
\end{equation}
In the remaining text we are assuming that $|C\left(P,Q\right)|=c$.

However, we must consider cases with only partial
matches. Therefore, it is desirable a trimmed
approach that discards a percentage of the worst matches \cite{Chetverikov}. So, we sort the pairs of the set $C\left(P,Q\right)$ such that $d\left(\mathbf{x}_{i_{1}},\mathbf{y}_{i_{1}}\right) \leq d\left(\mathbf{x}_{i_{2}},\mathbf{y}_{i_{2}}\right) \leq \cdot\cdot\cdot \leq d\left(\mathbf{x}_{i_{c}},\mathbf{y}_{i_{c}}\right)$ and consider a trimming parameter $0 \leq \tau \leq 1$ and a trimming boolean function:
\begin{align}
f^{trim}(\mathbf{p},\mathbf{q},\tau)=\begin{cases}
1 & if\; d\left(\mathbf{p},\mathbf{q}\right) \leq d\left(\mathbf{x}_{i_{c(1-\tau)}},\mathbf{y}_{i_{c(1-\tau)}}\right) \\
0 & otherwise
\end{cases}.
\label{eq:trimming-boolean-func}
\end{align}

So, we can build a new correspondence relation as:
\begin{equation}\label{eq:matching00-1}
C_{1}\left(P,Q,\tau \right)=\left\{\left(\mathbf{x}_{i},\mathbf{y}_{i}\right)\in C\left(P,Q\right); \ f^{trim}(\mathbf{x}_{i},\mathbf{y}_{i},\tau)=1 \right\},
\end{equation}
which is supposed to have $|C_{1}\left(P,Q,\tau \right)|=n$. We must notice that $C_{1}\left(P,Q,\tau \right) = C\left(P,Q\right)$ if $\tau = 0$.

The relationship defined by the expression (\ref{eq:matching00-1}) is based
on the distance function and nearest neighbor computation. We could
also consider shape descriptors computed over each point cloud. Generally
speaking, given a point cloud $S$, the shape descriptors can be
formulated as a function $f:S\rightarrow\mathbb{P}\left(\mathbb{R}\right)$,
where $\mathbb{P}\left(\mathbb{R}\right)$is the set of all subsets
of $\mathbb{R}$, named the power set of $\mathbb{R}$. In
this case, besides the distance criterion, we can also include shape
information in the correspondence computation by applying a boolean
correspondence function $f^{c}:P\times Q\rightarrow\left\{ 0,1\right\} $
such that \cite{Diez}:
\begin{align}
f^{c}(\mathbf{p},\mathbf{q})=\begin{cases}
1, & if\; f\left(\mathbf{p}\right)\approx f\left(\mathbf{q}\right),\\
0, & otherwise.
\end{cases}\label{eq:group-springs-02}
\end{align}

Also, before building $C\left(P,Q\right)$ in expression (\ref{eq:matching00}) we could perform a down-sampling
in the two point sets, based on the selection of key points through
the shape function, or through a naive interlaced sampling over same
spatial data structure \cite{LiXue11}.

Anyway, at the end of the matching process we get a base of the
set $P$, denoted by $X=\left\{ \mathbf{x}_{1},\mathbf{x}_{2},\ldots,\mathbf{x}_{n}\right\} \subset P$,
and a base of the set $Q$, denoted by $Y=\left\{ \mathbf{y}_{1},\mathbf{y}_{2},\ldots,\mathbf{y}_{n}\right\} \subset Q$
such that $C_{1}\left(P,Q,\tau \right)$ stands for the set of $n$ correspondence
pairs $\left(\mathbf{x}_{i},\mathbf{y}_{i}\right)\in X\times Y$.
This matching relation will be used to estimate a rigid transformation
that aligns the point clouds $P$ and $Q$. Specifically, we seek
for a rotation matrix $R$ and a translation $\mathbf{t}$ that minimizes
the mean squared error:
\begin{equation}
e^{2}\left(R,\mathbf{t}\right)=\frac{1}{n}\sum_{i=1}^{n}\left\Vert \mathbf{y}_{i}-\left(R\mathbf{x}_{i}+\mathbf{t}\right)\right\Vert _{2}^{2},\label{eq:eq2}
\end{equation}
which is used as a measure of the distance between the target set
$Q$ and the transformed source point cloud $\rho\left(P\right)=\left\{ \rho\left(\mathbf{p}_{1}\right),\rho\left(\mathbf{p}_{2}\right),\ldots,\rho\left(\mathbf{p}_{n}\right)\right\} $,
with $\rho$ defined by equation (\ref{eq:rigid-transf}). Now, we focus
in the specific three-dimensional case ($m=3$) and state the fundamental
theorem that steers most of the solutions for the registration
problem in $\mathbb{R}^{3}$.
\begin{theorem}
Let $X=\left\{ \mathbf{x}_{1},\mathbf{x}_{2},\ldots,\mathbf{x}_{n}\right\} \subset\mathbb{R}^{3}$
and $Y=\left\{ \mathbf{y}_{1},\mathbf{y}_{2},\ldots,\mathbf{y}_{n}\right\} \subset\mathbb{R}^{3}$,
the centers of mass $\mathbf{\mu}_{x}$, $\mathbf{\mu}_{y}$ for the respective point sets $X$
and $Y$, the cross-covariance $\Sigma_{xy}$, and the matrices $A$
and $M$, given by:
\begin{equation}
\mathbf{\mu}_{x}=\frac{1}{n}\sum_{i=1}^{n} \mathbf{x}_{i},\label{eq:mean-x}
\end{equation}
\begin{equation}
\mathbf{\mu}_{y}=\frac{1}{n}\sum_{i=1}^{n} \mathbf{y}_{i},\label{eq:mean-y}
\end{equation}
\begin{equation}
\Sigma_{xy}=\frac{1}{n}\sum_{i=1}^{n}\left(\mathbf{y}_{i}- \mathbf{\mu}_{y}\right)\left(\mathbf{x}_{i}-\mathbf{\mu}_{x}\right)^{T}.\label{eq:covm}
\end{equation}
\begin{equation}
A=\left(\Sigma_{xy}-\Sigma_{xy}^{T}\right),\label{eq:anti-symmetric}
\end{equation}
\begin{equation}
M\left(\Sigma_{xy}\right)=\left[\begin{array}{cccc}
tr\left(\Sigma_{xy}\right) & A_{23} & A_{31} & A_{12}\\
A_{23}\\
A_{31} &  & \Sigma_{xy}+\Sigma_{xy}^{T}-tr\left(\Sigma_{xy}\right)I_{3}\\
A_{12}
\end{array}\right].\label{eq:matrix-M}
\end{equation}

Hence, the optimum rotation $R$ and translation \textbf{t}
vector that minimizes the error in expression (\ref{eq:eq2}) are
determined uniquely as follows \cite{Horn}. The matrix $R$ is computed through the unit
eigenvector $\mathbf{v}=\left(\begin{array}{cccc}
v_{0} & v_{1} & v_{2} & v_{3}\end{array}\right)^{T}$ of $M$, corresponding to its maximum eigenvalue:
\begin{equation}
R=\left[\begin{array}{ccc}
1 - 2(v^2_1 + v^2_2) & 2(v_0v_1-v_2v_3) & 2(v_0v_2+v_1v_3)\\
2(v_0v_1+v_2v_3) & 1 - (v^2_0 + v^2_2) & 2(v_1v_2-v_0v_3)\\
2(v_0v_2 - v_1v_3) & 2(v_2v_2+v_0v_3) & 1 - (v^2_0+v^2_1)\\
\end{array}\right],\label{eq:R-computed-by-quaternion}
\end{equation}
and $t$ is calculated through $R$ and centroids in expressions (\ref{eq:mean-x})-(\ref{eq:mean-y})
as:
\begin{equation}
\mathbf{t}=\mathbf{\mu}_{y}-R\mathbf{\mu}_{x}.\label{eq:sol01}
\end{equation}
\end{theorem}

\subsection{Iterative Closest Point \label{sec:ICP}}

The classical ICP \cite{Besl}, described in the Algorithm \ref{alg:ICP}, receives the source $P$ and target $Q$ point clouds and it is composed by two major steps, iterated until a stopping criterion is satisfied, usually the number of iterations or error threshold.

\begin{algorithm}[H]
\label{alg:ICP}
\SetAlgoLined
\KwData{$P=\left\{\mathbf{p}_{i}\in\mathbb{R}^{3}; \mathbf{p}_{i}=\left(p_{i_{1}},p_{i_{2}},p_{i_{3}}\right)^{T}\right\}$,
$Q=\left\{ \mathbf{q}_{i}\in\mathbb{R}^{3}; \mathbf{q}_{i}=\left(q_{i_{1}},q_{i_{2}},q_{i_{3}}\right)^{T}\right\}$; trimming
$\tau$;}
\Begin{
$P_{0}=P$, $s=0$.

$\varepsilon_0 = \infty$.

$R_0 = I_3$, $\mathbf{t}_0 = (0, 0, 0)^T$.

\Repeat{$\varepsilon_s > \varepsilon_{s-1}$}
{
Apply the transformation to all points of the source:
$P_{s+1}=R_{s}P_{s}+\mathbf{t}_s\equiv\left\{ R_{s}\mathbf{p}+\mathbf{t}_s,\quad\mathbf{p}\in P_{s}\right\} $.

Compute the matching relation $C_{1}\left(P_{s+1},Q,\tau \right)$ through
expression (\ref{eq:matching00}).

Compute the principal eigenvector $\mathbf{v}$ of the matrix
$M$ defined in (\ref{eq:matrix-M}).

Calculate the rotation matrix $R_{s+1}$ and translation
vector $\mathbf{t}_{s+1}$ using expressions (\ref{eq:R-computed-by-quaternion})-(\ref{eq:sol01}).

Compute the error between the two point sets: $\varepsilon_{s+1} = e^{2}\left(R_{s+1},\mathbf{t}_{s+1}\right)$, from (\ref{eq:eq2}).

$s\leftarrow s+1$.
}
\Return $R_s, \mathbf{t}$.
}
\caption{Iterative Closest Point}
\end{algorithm}

\subsection{ICP-CTSF \label{sec:icp-ctsf}}

The ICP-CTSF \cite{Cejnog:2017} implements a matching strategy using a feature invariant to rigid transformations, based on the shape of second-order orientation tensors associated to each point. A voting algorithm is used, divided into an isotropic and an anisotropic voting field. So, given a cloud point $\mathbf{p}\in P$, let $L_{k}(\mathbf{p})\subset P$ be the set of $k\%$ nearest neighbor of $\mathbf{p}$ and $\mathbf{s}\in L_{k}(\mathbf{p})$. We can define $\mathbf{v}_{ps} = \left(\mathbf{s}-\mathbf{p}\right)$, $\widehat{\mathbf{v}}_{ps}=\mathbf{v}_{ps}/||\mathbf{v}_{ps}||_{2}$, as well as the function:
\begin{equation}
\sigma\left(\mathbf{p}\right)=\sqrt{\frac{||\mathbf{s}_{f}-\mathbf{p}||_{2}^{2}}{\ln0.01}}.\label{eq:sigma}
\end{equation}
where $\mathbf{s}_{f}$ is farthest neighbor of $\mathbf{p}$, which
has influence $0.01$. Given these elements, we can compute the second-order
tensor field:
\begin{equation}
\mathbf{T}\left(\mathbf{p}\right)=\sum\limits _{\mathbf{s}\in L_{k}(\mathbf{p})}\exp\left[\dfrac{-||\mathbf{v}_{ps}||_{2}^{2}}{\sigma^{2}\left(\mathbf{p}\right)}\right]\cdot\left(\widehat{\mathbf{v}}_{ps}\cdot\widehat{\mathbf{v}}_{ps}^{T}\right),\label{eq:stage1}
\end{equation}
which is the isotropic voting field computed through a weighted sum
of tensors $\widehat{\mathbf{v}}_{ps}\cdot\widehat{\mathbf{v}}_{ps}^{T}$,
built from the function (\ref{eq:sigma}) and from the vote vectors
$\mathbf{v}_{ps}$, $\mathbf{s}\in L_{k}(\mathbf{p})$.

\begin{figure}[!t]
\begin{minipage}[t]{1\linewidth}%
\centering %
\mbox{%
\includegraphics[width=0.6\linewidth]{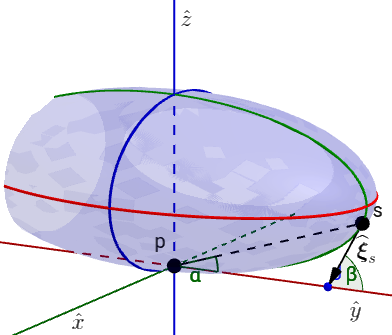}%
} %
\end{minipage}\protect\protect\caption{Geometric representation of the angles $\phi$, $\beta$ and unitary vector $\widehat{\mathbf{\xi}}_{s}$ of an arbitrary point $\mathbf{s}$.}
\label{fig:cjnog-geometry}
\end{figure}

Let the orthonormal basis generated by the eigenvectors $(\mathbf{e}_{1}\left(\mathbf{p}\right),\mathbf{e}_{2}\left(\mathbf{p}\right),\mathbf{e}_{3}\left(\mathbf{p}\right))$
of $\mathbf{T}\left(\mathbf{p}\right)$ and the corresponding eigenvalues
supposed to satisfy $\lambda_{3}\left(\mathbf{p}\right)<\lambda_{2}\left(\mathbf{p}\right)\leq\lambda_{1}\left(\mathbf{p}\right)$.
In this case, the local geometry at the point $\mathbf{p}$ can be
represented by the Figure \ref{fig:cjnog-geometry} where we picture
together the following elements: the coordinate system $\widehat{x},\widehat{y},\widehat{z}$
define by the eigenvectors $(\mathbf{e}_{1}\left(\mathbf{p}\right),\mathbf{e}_{2}\left(\mathbf{p}\right),\mathbf{e}_{3}\left(\mathbf{p}\right))$,
the plane $\pi$ that contains the point $\mathbf{p}$ and the axis $\widehat{z}$, its neighbor $\mathbf{s}$,
and the ellipse $E\in\pi$ that is tangent to the $\widehat{x},\widehat{y}$
plane in $\mathbf{p}$ and is centered at $z\in\widehat{z}$. The
vector $\widehat{\mathbf{\xi}}_{s}$, that is unitary and tangent
to $E$ at $\mathbf{s}$, gives a way to build a different structuring
element that enhances coplanar structures in the sense that the angle
$\beta\thickapprox0$ if $\mathbf{s}$ is close to the $\widehat{x},\widehat{y}$
plane. Specifically, if $d_{e}(\mathbf{p},\mathbf{s})$ is the arc length from $\mathbf{p}$
to $\mathbf{s}$ as indicated in the Figure \ref{fig:cjnog-geometry},
we define a new weighting function:
\begin{equation}
g\left(\mathbf{p},\mathbf{s}\right)=\begin{cases}
\exp\left[\dfrac{-d_{e}(\mathbf{p},\mathbf{s})}{\sigma^{2}\left(\mathbf{p}\right)}\right],\quad\tan{\phi_{s}}\leq\tan{\phi_{max}},\\
0.0\hfill\hfill\hfill,\tan{\phi_{s}}>\tan{\phi_{max}},
\end{cases}
\label{eq:def-gps}
\end{equation}
where $\sigma^{2}\left(\mathbf{p}\right)$ is calculated by expression (\ref{eq:sigma}), $\phi_{s}$ is the angle between $\mathbf{v}_{ps}=\left(\mathbf{s}-\mathbf{p}\right)$ and the $\hat{x},\hat{y}$ plane, $\phi_{max}$ constrains the influence of points misaligned to the $\hat{x},\hat{y}$ plane, with $45^\circ$ an ideal choice, as a mid term between smoother
results and robustness to outliers \cite{Vieira}.
%Figure \ref{fig:angle} shows a cut of the plan $y'=0$ depicting the vector $\widehat{v}_{pq}$ and the angles $\phi_{q'}$ and $\beta_{q'}$ for an arbitrary point, in a case where $\alpha_{ellip} = 30^{\circ}$.

With the above elements in mind, it is defined the tensor field $\mathbf{S}\left(\mathbf{p}\right)$, that is composed by the weighted sum of the tensors built from the votes received on the point, with weights computed by expression (\ref{eq:def-gps}) for all the points that have $\mathbf{p}$ as a neighbor:
\begin{equation}
\mathbf{S}\left(\mathbf{p}\right)=\sum\limits _{\mathbf{s}\in L_{k}(\mathbf{p})}g\left(\mathbf{p},\mathbf{s}\right)\cdot\left(\widehat{\mathbf{\xi}}_{s}\cdot\widehat{\mathbf{\xi}}_{s}^{T}\right)\label{eq:essepe}
\end{equation}

The tensor field in expression (\ref{eq:essepe}) can be seen as a shape function $\mathbf{S}:P\rightarrow\mathbb{R}^{3\times3}$ whose descriptors at a point $\mathbf{p}\in P$ are the eigenvalues $\lambda_{i}^{\mathbf{S}}\left(\mathbf{p}\right), i=1,2,3.$ Therefore, given two points $\mathbf{p},\mathbf{q}$ such that $\mathbf{p}\in P$ and $\mathbf{q}\in Q$, we compare the corresponding (local) geometries using the comparative tensor shape factor (CTSF), defined as:
\begin{equation}
CTSF\left(\mathbf{p},\mathbf{q}\right)=\sum\limits _{i=1}^{3}\left(\lambda_{i}^{\mathbf{S}_1}\left(\mathbf{p}\right)-\lambda_{i}^{\mathbf{S}_2}\left(\mathbf{q}\right)\right)^{2},\label{eq:CTFS}
\end{equation}
where $\mathbf{S}_{1}:P\rightarrow\mathbb{R}^{3\times3}$ and $\mathbf{S}_{2}:Q\rightarrow\mathbb{R}^{3\times3}$
are tensors computed following expression (\ref{eq:essepe})
and $\lambda_{i}^{\mathbf{S}_{1}}\left(\mathbf{p}\right)$ and $\lambda_{i}^{\mathbf{S}_{2}}\left(\mathbf{q}\right)$
are the $i$th eigenvalues calculated in the points $\mathbf{p}\in P$
and $\mathbf{q}\in Q$, respectively.

The CTSF is used side by side with the Euclidean distance to produce
a correspondence set that takes into account not only the nearest
point (like in expression (\ref{eq:matching00})) but also the shape
information:%. This is implemented by using the boolean function
%\begin{align}
%f^{c}(\mathbf{p},\mathbf{q})=\begin{cases}
%1 & if\;||\mathbf{p}-\mathbf{q}||_{2}+w_{n}\cdot CTSF\left(\mathbf{p},\mathbf{q}\right),\\
%0 & otherwise,
%\end{cases}.\label{eq:group-springs-02-1}
%\end{align}
\begin{equation}
d_{c,m}(\mathbf{p},\mathbf{q},m)=||\mathbf{p}-\mathbf{q}||_{2}+w_{m}\cdot CTSF\left(\mathbf{p},\mathbf{q}\right), \label{eq:dcm}
\end{equation}
where $CTSF\left(\mathbf{p},\mathbf{q}\right)$ is given by Equation (\ref{eq:CTFS}), $w_{m}=w_{0}b^{m}$, with $b<1$, and $0<w_{m}<w_{0}$.

The parameter $w_{0}$ is the initial weight given to the CTSF and
$b$ controls the update size of the weighting factor. To avoid numerical
instabilities we set $w_{m}=0$, when $w_{m}\approx0$. This weighting
strategy is responsible for its coarse-to-fine behavior when inserted
in the matching step of the ICP algorithm, given by expression (\ref{eq:matching00}). Specifically, the ICP-CTSF procedure (Algorithm \ref{alg:ICP-CTSF}) calculates the correspondence relation:
\begin{equation}\label{eq:matching00-ctsf}
C_{2}\left(P,Q,m \right)=\left\{\left(\mathbf{x}_{i_{l}},\mathbf{y}_{i_{l}}\right)\in P\times Q; \
\forall\mathbf{y}_{i_{k}}\in Q, \
d_{c,m}\left(\mathbf{x}_{i_{l}},\mathbf{y}_{i_{k}},m\right)\geq d_{c,m}\left(\mathbf{x}_{i_{l}},\mathbf{y}_{i_{l}},m\right)  \right\} ,
\end{equation}
and uses it to define the set:
\begin{equation}\label{eq:matching00-CTFS-Trim}
C_{3}\left(P,Q,\tau, m \right)=\left\{\left(\mathbf{x}_{i},\mathbf{y}_{i}\right)\in C_{2}\left(P,Q,m \right); \ f^{trim}(\mathbf{x}_{i},\mathbf{y}_{i},\tau)=1 \right\},
\end{equation}
which is the correspondence set applied by the ICP-CTSF technique, which is summarized in the Algorithm \ref{alg:ICP-CTSF}.

\begin{algorithm}[H]
\label{alg:ICP-CTSF}
\SetAlgoLined
\KwData{$P=\left\{\mathbf{p}_{i}\in\mathbb{R}^{3}; \mathbf{p}_{i}=\left(p_{i_{1}},p_{i_{2}},p_{i_{3}}\right)^{T}\right\}$,
$Q=\left\{ \mathbf{q}_{i}\in\mathbb{R}^{3}; \mathbf{q}_{i}=\left(q_{i_{1}},q_{i_{2}},q_{i_{3}}\right)^{T}\right\}$; trimming
$\tau$; $b$, such that $0<b<1$; $w_{0}\gg 0$;}
\Begin{
$P_{0}=P$, $s=0$, $m=1$.

$\varepsilon_0 = \infty$.

$R_0 = I_3$, $\mathbf{t}_0 = (0, 0, 0)^T$.

\Repeat{$\varepsilon_{s} > \varepsilon_{s-1}$}
{
Apply the transformation to all points of the source:
$P_{s+1}=R_{s}P_{s}+\mathbf{t}_s\equiv\left\{ R_{s}\mathbf{p}+\mathbf{t}_s,\quad\mathbf{p}\in P_{s}\right\} $.

Compute the matching relation $C_{3}\left(P_{s+1},Q,\tau, m \right)$ through
expression (\ref{eq:matching00-CTFS-Trim}).

Compute the principal eigenvector $\mathbf{v}$ of the matrix
$M$ defined in (\ref{eq:matrix-M}).

Calculate the matrix rotation matrix $R_{s+1}$ and translation
vector $\mathbf{t}_{s+1}$ using expressions (\ref{eq:R-computed-by-quaternion})-(\ref{eq:sol01}).

Compute the error between the two point sets: $\varepsilon_{s+1} = e^{2}\left(R_{s+1},\mathbf{t}_{s+1}\right)$, from (\ref{eq:eq2}).

\If{$\varepsilon_{s+1} > \varepsilon_{s}$}
{$m \leftarrow m+1$.

$w_{m} \leftarrow w_{0}b^{m}$.
}

$s\leftarrow s+1$.
}
\Return $R_s, \mathbf{t}$.
}
\caption{ICP-CTSF Procedure}
\end{algorithm}

\subsection{SWC-ICP Technique}

In this technique, besides the correspondence relation (\ref{eq:matching00}), we also use the correspondence set:
\begin{equation}
C_{CTSF}\left(P,Q\right)=\{\left(\mathbf{s}_{i},\mathbf{y}_{i}\right)\in P\times Q;~\mathbf{s}_{i}=\arg\min_{\mathbf{p}\in P}\left(CTSF(\mathbf{p},\mathbf{y}_{i})\right)\}, \label{eq:ctsf-selection}
\end{equation}
which contains the pairs of points $\left(\mathbf{p}_{i},\mathbf{s}_{i}\right)\in P\times Q$
whose local shapes are the most similar, according to the $CTSF$
criterion calculated by expression (\ref{eq:CTFS}). In order to combine
both the correspondence sets, we firstly develop expression (\ref{eq:covm})
to get:
\begin{equation}
\Sigma_{xy}=\frac{1}{n}\sum_{i=1}^{n}\left(\mathbf{y}_{i}\mathbf{x}_{i}^{T}\right)-\mathbf{\mu}_{y}\mathbf{\mu}_{x}^{T}.\label{eq:cov-new-rep}
\end{equation}

So, if we take expression (\ref{eq:eq2}) and perform the substitution:
\begin{equation}
\mathbf{x}_{i}\leftarrow\mathbf{x}_{i}+\omega_{n}\mathbf{s}_{i}\label{eq:change00}
\end{equation}
with $\omega_{n}\in\mathbb{R}$, we can write the mean
squared error (\ref{eq:eq2}) as:
\begin{equation}
e^{2}\left(R,\mathbf{t}\right)=\frac{1}{n}\sum_{i=1}^{n}\left\Vert \mathbf{y}_{i}-\left[R\left(\mathbf{x}_{i}+\omega_{n}\mathbf{s}_{i}\right)+\mathbf{t}\right]\right\Vert _{2}^{2}.\label{eq:eq2-1}
\end{equation}

Also, by substituting the variable change (\ref{eq:change00}) in
expression (\ref{eq:mean-x}) we get:
\begin{equation}
\mu_{x+\omega_{n}\mathbf{s}}=\frac{1}{n}\sum_{i=1}^{n}\left(\mathbf{x}_{i}+\omega_{n}\mathbf{s}_{i}\right)=\left(\frac{1}{n}\sum_{i=1}^{n}\mathbf{x}_{i}\right)+\omega_{n}\left(\frac{1}{n}\sum_{i=1}^{n}\mathbf{s}_{i}\right)\equiv\mathbf{\mu}_{x}+\omega_{n}\mathbf{\mu}_{s},\label{eq:mean-x-1}
\end{equation}
and, consequently:
\begin{equation}
\Sigma_{x+\omega_{n}\mathbf{s},y}=\frac{1}{n}\sum_{i=1}^{n}\left[\mathbf{y}_{i}\left(\mathbf{x}_{i}+\omega_{n}\mathbf{s}_{i}\right)^{T}\right]-\mathbf{\mu}_{y}\left(\mathbf{\mu}_{x}+\omega_{n}\mathbf{\mu}_{s}\right)^{T},\label{eq:covm-1}
\end{equation}
where $\mu_{y}$ is computed by equation (\ref{eq:mean-y}). We shall notice that the matrix (\ref{eq:covm-1}) combines the matching relations (\ref{eq:matching00}) and (\ref{eq:ctsf-selection}) being fundamental for the SWC-ICP described in \cite{Yamada}. According to the Theorem 1, the optimum rotation matrix $R$ and translation vector \textbf{t} that minimizes the error in expression (\ref{eq:eq2-1}) are uniquely determined by equations (\ref{eq:R-computed-by-quaternion})-(\ref{eq:sol01}) where $\mathbf{v}=(\begin{array}{cccc}
v_{0} & v_{1} & v_{2} & v_{3}\end{array})^{T}$ is the unit eigenvector of $M\left(\Sigma_{x+\omega_{n}\mathbf{s},y}\right)$ corresponding to the maximum eigenvalue. However, the SWC-ICP methodology achieves a coarse-to-fine behavior through the use of the weighting strategy of the ICP-CTSF. The SWC-ICP technique can be summarized in the Algorithm \ref{alg:SWC-ICP}.

\begin{algorithm}[H]
\label{alg:SWC-ICP}
\SetAlgoLined
\KwData{$P=\left\{\mathbf{p}_{i}\in\mathbb{R}^{3}; \mathbf{p}_{i}=\left(p_{i_{1}},p_{i_{2}},p_{i_{3}}\right)^{T}\right\}$,
$Q=\left\{ \mathbf{q}_{i}\in\mathbb{R}^{3}; \mathbf{q}_{i}=\left(q_{i_{1}},q_{i_{2}},q_{i_{3}}\right)^{T}\right\}$; trimming $\tau$; $b$, such that $0<b<1$; $w_{0}\gg 0$;}
\Begin{
$P_{0}=P$, $s=0$, $m=1$.

$\varepsilon_0 = \infty$.

$R_0 = I_3$, $\mathbf{t}_0 = (0, 0, 0)^T$.

Compute the matching relations $C_{CTSF}\left(P_{j},Q\right)$ through expression (\ref{eq:ctsf-selection}).

%Build the covariance matrix from (\ref{eq:covm-1}) using the shape correspondences.

\Repeat{$\varepsilon_{s} > \varepsilon_{s-1}$}
{
Apply the transformation to all points of the source:
$P_{s+1}=R_{s}P_{s}+\mathbf{t}_s\equiv\left\{ R_{s}\mathbf{p}+\mathbf{t}_s,\quad\mathbf{p}\in P_{s}\right\} $.

Compute the matching relation $C_{1}\left(P_{s+1},Q,\tau \right)$ through
expression (\ref{eq:matching00}).

Build the covariance matrix from (\ref{eq:covm-1}) using the shape correspondences (\ref{eq:ctsf-selection}) and the nearest neighbors (\ref{eq:matching00}).

Compute the matrix $M$ (\ref{eq:matrix-M}) using (\ref{eq:covm-1}) instead of (\ref{eq:covm}).

Compute the principal eigenvector $\mathbf{v}$ of the matrix
$M$.

Calculate the matrix rotation matrix $R_{s+1}$ and translation
vector $\mathbf{t}_{s+1}$ using expressions (\ref{eq:R-computed-by-quaternion})-(\ref{eq:sol01}).

Compute the error between the two point sets: $\varepsilon_{s+1} = e^{2}\left(R_{s+1},\mathbf{t}_{s+1}\right)$, from (\ref{eq:eq2}).

\If{$\varepsilon_{s+1} > \varepsilon_{s}$}
{$m \leftarrow m+1$.

$w_{m} \leftarrow w_{0}b^{m}$.
}

$s\leftarrow s+1$.
}
\Return $R_s, \mathbf{t}$.
}
\caption{SWC-ICP Technique}
\end{algorithm}

\section{Lie Groups and Lie Algebras \label{sec:Lie}}

In this section we present some results and concepts of Lie groups
theory that are relevant to problems in computer vision. We start
with the tensor product operation which allows to define tensors and
tensor fields. Next, we present the topological space definition which
relies only upon set theory. This concept gives the support to define
a differentiable manifold. The latter when integrated with the algebraic
structure of groups yields the framework of Lie groups.

\subsection{Tensor Product Spaces \label{subsec:Tensor-Product}}

The simplest way to define the tensor product of two vector spaces
$V_{1}$ and $V_{2}$, with dimensions $dim\left(V_{1}\right)=n$
and $dim\left(V_{2}\right)=m$, is by creating new vector spaces analogously
to multiplication of integers. For instance if $\{\mathbf{e}_{1}^{1},\mathbf{e}_{1}^{2},\mathbf{e}_{1}^{3},...,\mathbf{e}_{1}^{n}\}$
and $\{\mathbf{e}_{2}^{1},\mathbf{e}_{2}^{2},\mathbf{e}_{2}^{3},...,\mathbf{e}_{2}^{m}\}$
are basis in $V_{1}$ and $V_{2}$, respectively, then, the tensor
product between these spaces, denoted by $V_{1}\mathbf{\otimes}V_{2}$,
is a vector space that has the following properties:
\begin{enumerate}
\item Dimension:
\begin{equation}
dim\left(V_{1}\mathbf{\otimes}V_{2}\right)=n.m,\label{tensor000}
\end{equation}
\item Basis:
\begin{equation}
V_{1}\mathbf{\otimes}V_{2}=span\left(\left\{ \mathbf{e}_{1}^{i}\otimes\mathbf{e}_{2}^{j};\quad1\leq i\leq n,\quad1\leq j\leq m\right\} \right),\label{eq:tensor-prod-basis}
\end{equation}
\item Tensor product of vectors (Multilinear): Given $\mathbf{v}=\sum_{i=1}^{n}v_{i}\mathbf{e}_{1}^{i}$
and $\mathbf{u}=\sum_{j=1}^{m}u_{j}\mathbf{e}_{2}^{j}$ we define:
\begin{equation}
\mathbf{v\otimes u}=\sum_{i=1}^{n}\sum_{j=1}^{m}v_{i}u_{j}\mathbf{e}_{1}^{i}\mathbf{\otimes e}_{2}^{j}.\label{entan00-1}
\end{equation}
\end{enumerate}
Generically, given vector spaces $V_{1},V_{2},\ldots,V_{n}$, such
that $dim\left(V_{i}\right)=m_{i}$, and $\{\mathbf{e}_{i}^{1},\mathbf{e}_{i}^{2},...,\mathbf{e}_{i}^{m_{i}}\}$
is a basis for $V_{i}$ , we can define:

\begin{equation}
V_{1}\mathbf{\otimes}V_{2}\otimes\ldots\otimes V_{n}=span\left(\left\{ \mathbf{e}_{1}^{i_{1}}\otimes\mathbf{e}_{2}^{i_{2}}\otimes\ldots\otimes\mathbf{e}_{n}^{i_{n}};\quad\mathbf{e}_{k}^{i_{k}}\in V_{k}\right\} \right),\label{eq:general-tensor-product}
\end{equation}
and the properties above can be generalized in a straightforward way.

\subsection{Topological Spaces \label{subsec:topSpace}}

Given a set $E,$ we denote by $P\left(E\right)$ the set of the \textit{parts
}of $E,$ which is given by:

\begin{equation}
P\left(E\right)=\left\{ A;\quad A\subset E\right\} \label{set7}
\end{equation}

A topology over $E$ is a set $\Theta\subset P\left(E\right)$ that
fulfills the following axioms \cite{Honig76}:\\
\\
e1) Let $I$ an index set. If $O_{i}$ $\in$ $\Theta,$ then the
union $O=\begin{array}[t]{c}
{\cup}\\
{i\in I}
\end{array}O_{i}$ satisfies $O\in\Theta;$ \\
\\
e2) $O_{1},O_{2}\in\Theta\Rightarrow$ $O=O_{1}\cap O_{2}$ is such
that $O\in\Theta;$ \\
\\
e3) $E\in\Theta.$ \\
\\
The pair $\left(E,\Theta\right)$ is called a \textit{topological space}.
The elements of $E$ are called points and the elements of $\Theta$
are called open sets. A subset $A\subset E$ is named closed if its
complement is open.

We must remember that:

\[
\begin{array}[t]{c}
{\cup}\\
{i\in\emptyset}
\end{array}O_{i}=\emptyset,
\]
and so, the empty set $\emptyset$ must belong to $\Theta.$

Given a point $\mathbf{p}\in E$ , we call a \textbf{neighborhood} of $\mathbf{p}$
any set $V_{p}$ that contains an open set $O$ such that $\mathbf{p}\in O.$
A topological space $\left(E,\Theta\right)$ is \textbf{separated}
or \textbf{Hausdorff} if every two distinct points have disjoint neighborhoods;
that is, if given any distinct points $\mathbf{p}\neq \mathbf{q},$ there are neighborhoods
$V_{p}$ and $V_{q}$ such that $V_{p}\cap V_{q}=\emptyset.$

It is possible to define the concept of continuity in topological
spaces. Given two topological spaces $E_{1},E_{2}$ and a function
$f:E_{1}\rightarrow E_{2}$ we say that $f$ is continuous in a point
$p\in E_{1}$ if $\forall V_{f\left(p\right)}$ there is a neighborhood
$V_{p}$ such that $f\left(V_{p}\right)\subset V_{f\left(p\right)}.$

\subsection{Differentiable Manifolds and Tensors \label{subsec:Differentiable-Manifolds}}

Let $\M$ be a Hausdorff space as defined in section \ref{subsec:topSpace}.
A differential structure of dimension $n$ on $\M$ is a family of
one-to-one functions $\varphi_{\alpha}:U_{\alpha}\rightarrow$ $\mathbb{R}^{n}$,
where $U_{\alpha}$ is an open set of $\M$ and $\varphi_{\alpha}\left(U_{\alpha}\right)$
is an open set of $\mathbb{R}^{n}$, such that:

1) $\begin{array}[t]{c}
{\cup}\\
{\alpha}
\end{array}U_{\alpha}= \M.$

2) For every $\alpha,\beta,$ the mapping $\varphi_{\beta}\circ\varphi_{\alpha}^{-1}:\varphi_{\alpha}\left(U_{\alpha}\cap U_{\beta}\right)\rightarrow\varphi_{\beta}\left(U_{\alpha}\cap U_{\beta}\right)$
is a differentiable function.

A differentiable manifold of dimension $n$, denoted by $\M^{n}$,
is a Hausdorff space with a differential structure of dimension $n$
for which the family $\left\{ \left(U_{\alpha},\varphi_{\alpha}\right)\right\} $
is maximal respect to properties (1) e (2). If we replace differentiable
to smooth in property (2) then $\M$ is called a smooth manifold.

The family $\left\{ \left(U_{\alpha},\varphi_{\alpha}\right)\right\} $
is an atlas on the topological space $\M$ and each pair $\left(U_{\alpha},\varphi_{\alpha}\right)$
is named a chart.

Let $\mathbf{p}\in U_{\alpha}$ and $\varphi_{\alpha}\left(\mathbf{p}\right)=\left(x_{1}\left(\mathbf{p}\right),...,x_{n}\left(\mathbf{p}\right)\right)\in\mathbb{R}^{n}$.
Then, the set $\varphi_{\alpha}\left(U_{\alpha}\right)$ gives a local
coordinate system in $\mathbf{p}$ and $x_{i}\left(p\right)$
are the local coordinates of $\mathbf{p}$.

Let $C$ be a smooth curve in the manifold $\M$, parameterized as
$\phi:I\subset \mathbb{R} \rightarrow \M.$ In local coordinates $\varphi_{\alpha}\left(U_{\alpha}\right)$
the curve $C$ is defined by $n$ smooth functions $\varphi_{\alpha}\circ\phi\left(\varepsilon\right)=\left(x_{1}\left(\varepsilon\right),...,x_{n}\left(\varepsilon\right)\right)$
of the real variable $\varepsilon.$ Let $\mathbf{p}=\phi\left(0\right)$
the point of $C$ for $\varepsilon=0$. Then, the tangent vector at
$C$ in $\mathbf{p}$ is given by the derivative:

\[
\mathbf{v}=\frac{d\varphi_{\alpha}\circ\phi\left(0\right)}{d\varepsilon}=\left(\frac{dx_{1}\left(0\right)}{d\varepsilon},...,\frac{dx_{n}\left(0\right)}{d\varepsilon}\right)\equiv\acute{\phi}\left(0\right).
\]

The tangent vector $\mathbf{v}$ to $\M$ at $\mathbf{p}$
can be expressed in the local coordinates $\mathbf{x}$ as:

\begin{equation}
\mathbf{v}=\sum_{i=1}^{n}\left(v_{i}\frac{\partial}{\partial x_{i}}\right).\label{tangent-vector-def}
\end{equation}
where $v_{i}=dx_{i}/d\varepsilon$ and the vectors $\partial/\partial x_{i}$
are defined by the local coordinates.

So, we can define the tangent space to a manifold $\M$ at a point $\mathbf{p}\in \M$
as:

\begin{equation}
T_{p}\M=\left\{ \acute{\phi}\left(0\right),\;with\;\phi:I\subset \mathbb{R} \rightarrow \M\;differentiable\;and\;\phi\left(0\right)=\mathbf{p}\right\} .\label{eq:al00}
\end{equation}

We can show that $T_{p}\M$ is a vector space, isomorphic to $\mathbb{R}^{n}$.
Besides, the set:

\begin{equation}
B=\left\{ \frac{\partial}{\partial x_{1}},...,\frac{\partial}{\partial x_{n}}\right\} ,\label{base-tangent-space}
\end{equation}
gives a natural basis for $T_{p}\M$.

The collection of all tangent spaces to $\M$ is the tangent bundle
of the differentiable manifold $\M:$

\begin{equation}
T\M=\bigcup_{\mathbf{p}\in\M}T_{\mathbf{p}}\left(\M\right) \equiv \left\{ \left(\mathbf{p},\mathbf{v}\right);\quad \mathbf{p} \in \M \quad and \quad \mathbf{v} \in T_{\mathbf{p}}\left(\M\right) \right\} .\label{lie3}
\end{equation}

A Riemannian manifold is a manifold $\M$ equipped with an inner product
in each point $\mathbf{p}$ (bilinear form in the tangent space
$T_{\mathbf{p}}\left(\M\right)$) that varies smoothly from point
to point. A geodesic in a Riemannian manifold $\M$ is a differentiable
curve $\alpha:I\subset\R\rightarrow\M$ that is the shortest path
between any two points $\mathbf{p}_{1}=\alpha\left(t_{1}\right)$ and $\mathbf{p}_{2}=\alpha\left(t_{2}\right)$(see
\cite{DiffManifoldsRiemann}).

A vector field $\mathbf{v}$ over a manifold $\M$ is a function that associates
to each point $\mathbf{p}\in\M$ a vector $\mathbf{v}(\mathbf{p})\in T_{p}\left(\M\right)$.
Therefore, in the local coordinates:
\begin{equation}
\mathbf{v}(\mathbf{p})=\sum_{i=1}^{m}\left(v_{i}(p)\frac{\partial}{\partial x_{i}}(p)\right),\label{vector-field}
\end{equation}
where now we explicit the fact that expressions (\ref{tangent-vector-def})
and (\ref{base-tangent-space}) are computed in each point $\mathbf{p}\in\M$.

The notion of tensor field is formulated as a generalization of the
vector field using the concept of tensor product of section
\ref{subsec:Tensor-Product}. So, given the subspaces $T_{p}^{i}\left(\M\right)\subset T_{p}\left(\M\right)$,
with $dim\left(T_{p}^{i}\left(\M\right)\right)=m_{i}$, $i=1,2,\cdot\cdot\cdot,n$,
the tensor product of these spaces, denoted by $T_{p}^{1}\left(\M\right)\mathbf{\otimes}T_{p}^{2}\left(\M\right)\mathbf{\otimes}\cdot\cdot\cdot\mathbf{\otimes}T_{p}^{n}\left(\M\right)$,
is a vector space defined by expression (\ref{eq:general-tensor-product})
with $V_{i}=T_{p}^{i}\left(\M\right)$ and individual basis $\left\{ \mathbf{e}_{k}^{i_{k}}(p),i_{k}=1,2,\cdot\cdot\cdot,m_{k}\right\} \subset T_{p}^{k}\left(\M\right)$;
that means, a natural basis $B$ for the vector space $T_{p}^{1}\left(\M\right)\mathbf{\otimes}T_{p}^{2}\left(\M\right)\mathbf{\otimes}\cdot\cdot\cdot\mathbf{\otimes}T_{p}^{n}\left(\M\right)$
is the set:
\begin{equation}
B=\left\{ \mathbf{e}_{1}^{i_{1}}(p)\mathbf{\otimes}\mathbf{e}_{2}^{i_{2}}(p)\mathbf{\otimes}\cdot\cdot\cdot\mathbf{\otimes}\mathbf{e}_{n}^{i_{n}}(p),\quad\mathbf{e}_{k}^{i_{k}}(p)\in T_{p}^{k}\left(\M\right)\right\} .\label{tensor-basis00}
\end{equation}

In this context, a tensor $\mathbf{X}$ of order $n$ in $p\in\M$
is defined as an element $\mathbf{X}\left(p\right)\in T_{p}^{1}\left(\M\right)\mathbf{\otimes}T_{p}^{2}\left(\M\right)\mathbf{\otimes}\cdot\cdot\cdot\mathbf{\otimes}T_{p}^{n}\left(\M\right);$
that is, an abstract algebraic entity that can be expressed as \cite{Liu-Ruan2011}:

\begin{equation}
\mathbf{X}\left(p\right)=\sum_{i_{1},i_{2},\cdot\cdot\cdot,i_{n}}^ {}\mathbf{X}_{i_{1},i_{2},\cdot\cdot\cdot,i_{n}}\left(p\right)\mathbf{e}_{1}^{i_{1}}\left(p\right)\mathbf{\otimes}\mathbf{e}_{2}^{i_{2}}\left(p\right)\mathbf{\otimes}\cdot\cdot\cdot\mathbf{\otimes}\mathbf{e}_{n}^{i_{n}}\left(p\right).\label{eq:tensor-field-manifold-1}
\end{equation}

Analogously to the vector case, a tensor field of order $n$ over
a manifold $\M$ is a function that associates to each point $\mathbf{p}\in\M$
a tensor $\mathbf{X}\left(p\right)\in T_{p}^{1}\left(\M\right)\mathbf{\otimes}T_{p}^{2}\left(\M\right)\mathbf{\otimes}\cdot\cdot\cdot\mathbf{\otimes}T_{p}^{n}\left(\M\right)$.
On the other hand, analogously to the tangent bundle, we can define the tensor bundle as:
\[
T\M\left(n\right)=\bigcup_{\mathbf{p}\in\M}T_{p}^{1}\left(\M\right)\mathbf{\otimes}T_{p}^{2}\left(\M\right)\mathbf{\otimes}\cdot\cdot\cdot\mathbf{\otimes}T_{p}^{n}\left(\M\right)
\]

\begin{equation}
\equiv \left\{ \left(p,\mathbf{X}\right);\;p\in\M\;and\;\mathbf{X}\in T_{p}^{1}\left(\M\right)\mathbf{\otimes}T_{p}^{2}\left(\M\right)\mathbf{\otimes}\cdot\cdot\cdot\mathbf{\otimes}T_{p}^{n}\left(\M\right)\right\}, \label{lie3-1}
\end{equation}
where we shall notice that $T\M\left(n\right)=T\M$ if $n=1$.

\subsection{Group Definition \label{subsec:group}}

A group is a pair $\left(G,\cdot\right)$ where $G$ is a non-empy
set and $\cdot:G\times G\longmapsto G$ is a mapping, also called
an operation \cite{Lang1984}, with the following properties:

(1) Associative. If $g,h,k\in G$, then:

\[
g\cdot\left(h\cdot k\right)=\left(g\cdot h\right)\cdot k.
\]

(2) Unity element. There is an element $e\in G$ , named unit element,
such that:

\[
e\cdot g=g\cdot e=g,
\]

for all $g\in G.$

(3) Inverse. For any $g\in G$ there exists an element $g^{-1}\in G$,
called inverse, that satisfies:
\[
g^{-1}\cdot g=g\cdot g^{-1}=e.
\]
$\blacksquare$

A commutative (or abelian) groups satisfies:
\begin{equation}
g\cdot h=h\cdot g,\quad\forall g,h\in G\label{eq:Abelian}
\end{equation}
\\
 \textbf{Examples of Abelian Groups:} $\left(\mathbb{Z},+\right)$ and $\left(\mathbb{R},+\right)$
, where '$+$' is the usual addition.\\

\textbf{Matrix Groups: }Let $M_{n}\left(\mathbb{R}\right)$ be the
real $n\times n$ matrices. We shall consider the usual matrix multiplication
$\cdot:M_{n}\left(\mathbb{R}\right)\times M_{n}\left(\mathbb{R}\right)\longrightarrow M_{n}\left(\mathbb{R}\right)$.
In this case, we can show that the following subsets of $M_{n}\left(\mathbb{R}\right)$
are groups regarding matrix multiplication.

2) $GL_{n}\left(\mathbb{R}\right)$: real $n\times n$ matrices with
non-null determinant

4) $PDUT\left(n\right)$ : the set of $n\times n$ upper triangular
matrices with positive diagonal entries

5) $SPD$ or $Sym^{+}\left(n\right)$: symmetric positive definite
matrices

7) $GL^{+}\left(n\right)$: the group of $n\times n$ matrices with
positive determinant

\subsection{Lie Groups and Lie Algebras}

An $r-parametric$ Lie group $G$ is a group and also a differentiable
manifold of dimension $r$ such that the group operation \cite{Olver86}:

\[
m:G\times G\longrightarrow G,\text{ }m\left(g,h\right)=g\cdot h,\text{ }g,h\in G,
\]
and inversion:

\[
i:G\rightarrow G,\text{ }i\left(g\right)=g^{-1},\text{ }g\in G,
\]
are smooth functions.

Examples:

(a) Let $G=\mathbb{R}^{r}$ with the trivial differential structure given
by the identity mapping and the usual addition operation in $\mathbb{R}^{r}$,
where the inverse of a vector $x$ is its opposite $-x.$ Once both the
addition and inversion are smooth mappings, $\mathbb{R}^{r}$ is an example
of a r-parametric Lie group.

(b) Consider the set $G=SO(2)$ composed by the rotations:
\[
SO\left(2\right)=\left\{ \left(\begin{array}{ll}
\cos\left(\theta\right) & -sen\left(\theta\right)\\
sen\left(\theta\right) & \cos\left(\theta\right)
\end{array}\right);\text{ }0\leq\theta<2\pi\right\} ,
\]
with the usual matrix multiplication. It is straightforward to show
that the group axioms are satisfied and, consequently, $SO(2)$ os
a group. Its differential structure can be obtained by the observation
that we can identify $SO(2)$ with the unitary circle \cite{Olver86}:

\[
S^{1}=\left\{ \left(\cos\left(\theta\right),sen\left(\theta\right)\right);\text{ }0\leq\theta<2\pi\right\},
\]
and, consequently, $SO\left(2\right)$ is a $r-parametric$ Lie group.

Let $\left(G,\cdot\right)$ and $\left(\widetilde{G},\circ\right)$ two
Lie groups. A homeomorphism between these groups is a smooth function $\varphi:G\rightarrow\widetilde{G}$
that satisfies:

\[
\varphi\left(a\cdot b\right)=\varphi\left(a\right)\circ\varphi\left(b\right),\quad\forall a,b\in G.
\]

Besides, if $\varphi$ is bijective and its inverse $\varphi^{-1}$ is
smooth, we say that $\varphi$ is an isomorphism between $\left(G,\cdot\right)$
and $\left(\widetilde{G},\circ\right)$.

If $G\subset GL_{n}\left(\mathbb{R}\right)$ then the tangent space at $p$ satisfies $T_{p}G\subset M_{n}\left(\mathbb{R}\right)$. Moreover,
the \textbf{Lie Algebra }$\mathbf{g}$ associated to the group
$\left(G,\cdot\right)$ is the vector space $\mathbf{g}=T_{I}G$. Hence, considering the definition (\ref{eq:al00}) of tangent space,
we see that for all $X\in\mathbf{g}$ we have:

\begin{equation}
X=\frac{d\phi\left(0\right)}{d\varepsilon},\label{eq:prop0}
\end{equation}

\begin{equation}
\phi\left(0\right)=I,\label{eq:prop01}
\end{equation}
where $\phi:I \subset \mathbb{R} \rightarrow G.$ This fact motivates the following
definition.

\textbf{Definition 1:} Let G be a matrix Lie group. The Lie algebra
of G, denoted as $\mathbf{g}$, is the set of all matrices $X$
such that $\exp\left(\varepsilon X\right)\in G$ for all real numbers
$\varepsilon$. Therefore:

\[
\phi\left(\varepsilon\right)=\exp\left(\varepsilon X\right),\quad\varepsilon\in\mathbb{R},
\]
is a curve in the group $\left(G,\cdot\right)$. It is straightforward
to show that this curve satisfies the properties (\ref{eq:prop0})-(\ref{eq:prop01}).
Besides the matrix exponential is a function $\exp:M_{n}\left(\mathbb{R}\right)\longrightarrow GL_{n}\left(\mathbb{R}\right)$ that
can be computed by the series:

\begin{equation}
\exp\left(A\right)=\sum_{n\geq0}\frac{1}{n!}A^{n}.\label{eq:expmatrix}
\end{equation}

For $A\in M_{n}\left(\mathbb{R}\right)$ and $r>0$, let:

\[
N_{M_{n}\left(\mathbb{R}\right)}\left(A;r\right)=\left\{ B\in M_{n}\left(\mathbb{R}\right);\quad\left\Vert B-A\right\Vert <r\right\} ,
\]
which is an open disc of radius $r$ in $M_{n}\left(\mathbb{R}\right)$.
So, we can show that the function $\log:N_{M_{n}\left(\mathbb{R}\right)}\left(I;1\right)\longrightarrow M_{n}\left(\mathbb{R}\right)$
\cite{Baker:2002}:

\begin{equation}
\log\left(A\right)=\sum_{n\geq1}\frac{\left(-1\right)^{n-1}}{n}\left(A-I\right)^{n},\label{eq:logmat00}
\end{equation}
is well-defined and we can prove that $\exp\left(\log\left(A\right)\right)=A$.

\section{Embedding Gaussians into Linear Spaces \label{sec:Embed}}

In this section we summarize some fundamental results presented in
\cite{Li2017LocalLM}. So, we consider the spaces:

\begin{enumerate}
\item $N\left(n\right)$: Space of Gaussians of dimension $n$. Each Gaussian distribution is denoted by $\mathcal{N}\left(\mathbf{\mu},\Sigma\right)$, where $\mathbf{\mu}$ is the mean vector and $\Sigma$ the covariance matrix

\item $Ut\left(n\right)$: The set of all $n\times n$ upper triangular
matrices

\item $Sym\left(n\right)$: symmetric $n\times n$ matrices
\end{enumerate}

\begin{theorem}
The set $N\left(n\right)$ is a Riemmanian manifold.

Dem: \cite{Li2017LocalLM}.
\end{theorem}

Let $\Sigma,\Sigma^{-1}\in SPD$ and the Cholesky decomposition $\Sigma^{-1}=LL^{T}$,
where $L$ is a lower triangular matrix with positive diagonal. Steered
by this decomposition, we can define the following operation in $N\left(n\right)$.

\[
\star:N\left(n\right)\times N\left(n\right)\rightarrow N\left(n\right),
\]

\begin{equation}
\mathcal{N}\left(\mathbf{\mu}_{1},\Sigma_{1}\right)\star\mathcal{N}\left(\mathbf{\mu}_{2},\Sigma_{2}\right)=\mathcal{N}\left(L_{1}^{-T}\mathbf{\mu}_{2}+\mathbf{\mu}_{1},\left(L_{1}L_{2}\right)^{-T}\left(L_{1}L_{2}\right)^{-1}\right)\label{eq:produto00}
\end{equation}

\begin{theorem}
Under the operation given by expression (\ref{eq:produto00}),
the set $N\left(n\right)$ is a Lie group, denoted by $\left(N\left(n\right),\star\right)$

Dem: \cite{stanford1981riemannian}.
\end{theorem}

Now, let us consider the Lie group $\left(A^{+}\left(n+1\right),\cdot\right)$:

\[
A^{+}\left(n+1\right)=\left\{ A_{\mu,Z}=\left(\begin{array}{cc}
Z & \mathbf{\mu}\\
0^{T} & 1
\end{array}\right)\in\mathbb{R}^{\left(n+1\right)\times\left(n+1\right)},\:Z\in PDUT\left(n\right),\mathbf{\mu}\in\mathbb{R}^{n}\right\} ,
\]
where the operation '$\cdot$' is the usual matrix multiplication.

\begin{theorem}
The function $\varphi:A^{+}\left(n+1\right)\rightarrow N\left(n\right)$
given by:

\[
\varphi\left(A_{\mu,L^{-T}}\right)=\mathcal{N}\left(\mathbf{\mu},\Sigma\right),
\]
where $\Sigma=L^{-T}L^{-1}$ and $L^{-T}\in PDUT\left(n\right)$,
is an isomorphism between the Lie groups $\left(A^{+}\left(n+1\right),\cdot\right)$
and $\left(N\left(n\right),\star\right)$ .

Dem: \cite{Li2017LocalLM}.
\end{theorem}

Now, we take the set:

\[
A\left(n+1\right)=\left\{ A_{\mathbf{t},X}=\left(\begin{array}{cc}
X & \mathbf{t}\\
0^{T} & 0
\end{array}\right);\quad X\in Ut\left(n\right),\quad\mathbf{t}\in\mathbb{R}^{n}\right\} ,
\]
which is the Lie algebra of the matrix group $A^{+}\left(n+1\right)$.

\begin{theorem}
The function $\psi:A\left(n+1\right)\rightarrow A^{+}\left(n+1\right)$
given by:

\[
\psi\left(A_{\mathbf{t},X}\right)=\exp\left(A_{\mathbf{t},X}\right),
\]
is a smooth bijection and its inverse is smooth as well.

Dem: \cite{Li2017LocalLM}.
\end{theorem}

\begin{theorem}
Log-Euclidean on $A^{+}\left(n+1\right)$): We
define the operation:

\[
\otimes:A^{+}\left(n+1\right)\times A^{+}\left(n+1\right)\longrightarrow A^{+}\left(n+1\right),
\]

\[
A_{1}\otimes A_{2}=\exp\left(\log\left(A_{1}\right)+\log\left(A_{2}\right)\right)
\]

and:

\[
\odot:\mathbb{R}\times A^{+}\left(n+1\right)\longrightarrow A^{+}\left(n+1\right),
\]

\[
\lambda\odot A=\exp\left(\lambda\log\left(A\right)\right)=A^{\lambda}.
\]

Under the operation $\otimes$, $A^{+}\left(n+1\right)$ is a commutative
Lie group. Besides:

\[
\log:A^{+}\left(n+1\right)\longrightarrow A\left(n+1\right),
\]

\[
A\mapsto\log\left(A\right),
\]
is a Lie group isomorphism. In addition, under $\otimes$ and $\odot$,
$A^{+}\left(n+1\right)$ is a linear space.

Dem: \cite{Li2017LocalLM}.
\end{theorem}

Consequently, we can complete an embedding process, denoted by DE-LogE
in \cite{Li2017LocalLM}, as:

\[
\mathcal{N}\left(\mathbf{\mu},\Sigma\right)\quad\underrightarrow{\varphi^{-1}}\quad A_{\mu,L^{-T}}\quad\underrightarrow{\log}\quad\log\left(A_{\mu,L^{-T}}\right)=\log\left(\begin{array}{cc}
L^{-T} & \mathbf{\mu}\\
0^{T} & 1
\end{array}\right),
\]
where $\Sigma=L^{-T}L^{-1}$ and $L^{-T}\in PDUT\left(n\right)$. Recall
that $L$ is the Cholesky factor of $\Sigma^{-1}$.

\section{Studying the DE-LogE \label{sec:DE-LogE}}

Let us denote:

\begin{equation}
A=\left(\begin{array}{cc}
L^{-T} & \mathbf{\mu}\\
0^{T} & 1
\end{array}\right).
\label{eq:matrixA}
\end{equation}

Therefore, according to expression (\ref{eq:logmat00}), if $\left\Vert A-I\right\Vert <1$
we can write:

\begin{equation}
\log\left(A\right)=\sum_{n\geq1}\frac{\left(-1\right)^{n-1}}{n}\left(A-I\right)^{n}.\label{eq:logmat00-1}
\end{equation}

Just as a prospective example, we are going to analyse the above Taylor
expansion for $n=5$:

\[
\sum_{n=1}^{5}\frac{\left(-1\right)^{n-1}}{n}\left(A-I\right)^{n}=\left(A-I\right)-\frac{1}{2}\left(A-I\right)^{2}+\frac{1}{3}\left(A-I\right)^{3}-\frac{1}{4}\left(A-I\right)^{4}+\frac{1}{5}\left(A-I\right)^{5}
\]

We shall notice that:

\[
A-I=\left(\begin{array}{cc}
L^{-T} & \mathbf{\mu}\\
0^{T} & 1
\end{array}\right)-\left(\begin{array}{cc}
I & 0\\
0^{T} & 1
\end{array}\right)=\left(\begin{array}{cc}
C & \mathbf{\mu}\\
0^{T} & 0
\end{array}\right),
\]
where $C=L^{-T}-I$.

Consequently:

\[
\sum_{n=1}^{5}\frac{\left(-1\right)^{n-1}}{n}\left(A-I\right)^{n}=\left(\begin{array}{cc}
T_{11}^{\left(5\right)} & T_{12}^{\left(5\right)}\\
0^{T} & 0
\end{array}\right),
\]
where:

\begin{equation}
T_{11}^{\left(5\right)}=C-\frac{1}{2}C^{2}+\frac{1}{3}C^{3}-\frac{1}{4}C^{4}+\frac{1}{5}C^{5}=\sum_{n=1}^{5}\frac{\left(-1\right)^{n-1}}{n}C^{n},\label{eq:T11}
\end{equation}

\begin{equation}
T_{12}^{\left(5\right)}=C^{-1}\left(C-\frac{1}{2}C^{2}+\frac{1}{3}C^{3}-\frac{1}{4}C^{4}+\frac{1}{5}C^{5}\right)\mathbf{\mu}=C^{-1}T_{11}\mu.\label{eq:T12}
\end{equation}

The pattern observed in equations (\ref{eq:T11})-(\ref{eq:T12})
is verified for all $n\geq1$. To demonstrate this, it is just a matter
to notice that:

\[
\left(\begin{array}{cc}
C & \mathbf{\mu}\\
0^{T} & 0
\end{array}\right)^{2}=\left(\begin{array}{cc}
C & \mathbf{\mu}\\
0^{T} & 0
\end{array}\right)\left(\begin{array}{cc}
C & \mathbf{\mu}\\
0^{T} & 0
\end{array}\right)=\left(\begin{array}{cc}
C^{2} & C\mathbf{\mu}\\
0^{T} & 0
\end{array}\right),
\]

\[
\left(\begin{array}{cc}
C & \mathbf{\mu}\\
0^{T} & 0
\end{array}\right)^{3}=\left(\begin{array}{cc}
C^{2} & C\mathbf{\mu}\\
0^{T} & 0
\end{array}\right)\left(\begin{array}{cc}
C & \mathbf{\mu}\\
0^{T} & 0
\end{array}\right)=\left(\begin{array}{cc}
C^{3} & C^{2}\mathbf{\mu}\\
0^{T} & 0
\end{array}\right),
\]

\[
\ldots
\]

\begin{equation}
\left(\begin{array}{cc}
C & \mathbf{\mu}\\
0^{T} & 0
\end{array}\right)^{k+1}=\left(\begin{array}{cc}
C^{k+1} & C^{k}\mathbf{\mu}\\
0^{T} & 0
\end{array}\right).\label{eq:powerCk1}
\end{equation}

Consequently, using an induction process, if we assume that:

\[
\sum_{n=1}^{k}\frac{\left(-1\right)^{n-1}}{n}\left(A-I\right)^{n}=\left(\begin{array}{cc}
T_{11}^{\left(k\right)} & T_{12}^{\left(k\right)}\\
0^{T} & 0
\end{array}\right),
\]
where:

\begin{equation}
T_{11}^{\left(k\right)}=\sum_{n=1}^{k}\frac{\left(-1\right)^{n-1}}{n}C^{n},\label{eq:T11-1}
\end{equation}

\begin{equation}
T_{12}^{\left(k\right)}=C^{-1}T_{11}^{\left(k\right)}\mathbf{\mu,}\label{eq:T12-1}
\end{equation}
then:

\[
\sum_{n=1}^{k+1}\frac{\left(-1\right)^{n-1}}{n}\left(A-I\right)^{n}=\left(\begin{array}{cc}
T_{11}^{\left(k\right)} & T_{12}^{\left(k\right)}\\
0^{T} & 0
\end{array}\right)+\frac{\left(-1\right)^{\left(k+1\right)-1}}{k+1}\left(\begin{array}{cc}
C & \mathbf{\mu}\\
0^{T} & 0
\end{array}\right)^{k+1}.
\]

Using equation (\ref{eq:powerCk1}) we obtain:

\[
\sum_{n=1}^{k+1}\frac{\left(-1\right)^{n-1}}{n}\left(A-I\right)^{n}=\left(\begin{array}{cc}
T_{11}^{\left(k\right)} & T_{12}^{\left(k\right)}\\
0^{T} & 0
\end{array}\right)+\frac{\left(-1\right)^{\left(k+1\right)-1}}{k+1}\left(\begin{array}{cc}
C^{k+1} & C^{k}\mathbf{\mu}\\
0^{T} & 0
\end{array}\right),
\]
and, consequently:

\[
T_{11}^{\left(k+1\right)}=\sum_{n=1}^{k}\frac{\left(-1\right)^{n-1}}{n}C^{n}+\frac{\left(-1\right)^{\left(k+1\right)-1}}{k+1}C^{k+1}=\sum_{n=1}^{k+1}\frac{\left(-1\right)^{n-1}}{n}C^{n},
\]

\[
T_{12}^{\left(k+1\right)}=C^{-1}T_{11}^{\left(k\right)}\mathbf{\mu}+\frac{\left(-1\right)^{\left(k+1\right)-1}}{k+1}C^{k}\mathbf{\mu}=C^{-1}\left(T_{11}^{\left(k\right)}+\frac{\left(-1\right)^{\left(k+1\right)-1}}{k+1}C^{k+1}\right)\mathbf{\mu}
\]

\[
=C^{-1}T_{11}^{\left(k+1\right)}\mathbf{\mu},
\]
which concludes the induction process.

Consequently, we can compute the function:

\[
LOG\left(C\right)=\sum_{n\geq1}\frac{\left(-1\right)^{n-1}}{n}C^{n},
\]
which is well defined if $\left\Vert C\right\Vert <1$ \cite{Baker:2002}, and
write:

\[
\log\left(A\right)=\sum_{n\geq1}\frac{\left(-1\right)^{n-1}}{n}\left(A-I\right)^{n}=\left(\begin{array}{cc}
T_{11} & T_{12}\\
0^{T} & 0
\end{array}\right),
\]
where

\begin{equation}
T_{11}=LOG\left(C\right)=\log\left(L^{-T}\right),\label{eq:LogE00}
\end{equation}

\begin{equation}
T_{12}=C^{-1}LOG\left(C\right)\mathbf{\mu}=\left(L^{-T}-I\right)^{-1}\log\left(L^{-T}\right)\mathbf{\mu}.\label{eq:LogE01}
\end{equation}

%Therefore, given $A_{1}$ and $A_{2}$:
%
%\[
%D=\log\left(A_{1}\right)-\log\left(A_{2}\right)=
%\]
%
%\[
%\left(\begin{array}{ccc}
%\log\left(L_{1}^{-T}\right) &  & \left(L_{1}^{-T}-I\right)^{-1}\log\left(L_{1}^{-T}\right)\mathbf{p}\\
%\\
%0^{T} &  & 0
%\end{array}\right)-\left(\begin{array}{ccc}
%\log\left(L_{2}^{-T}\right) &  & \left(L_{2}^{-T}-I\right)^{-1}\log\left(L_{2}^{-T}\right)\mathbf{q}\\
%\\
%0^{T} &  & 0
%\end{array}\right).
%\]
%
%So, the expressions:
%
%\begin{equation}
%D\left(1,1\right)=\log\left(L_{1}^{-T}\right)-\log\left(L_{2}^{-T}\right),\label{eq:D11}
%\end{equation}
%
%\begin{equation}
%D\left(1,2\right)=\left(L_{1}^{-T}-I\right)^{-1}\log\left(L_{1}^{-T}\right)\mathbf{p}-\left(L_{2}^{-T}-I\right)^{-1}\log\left(L_{2}^{-T}\right)\mathbf{q}, \label{eq:D12}
%\end{equation}
%are the fundamental ones for our proposal.

\section{Comparing Tensors in DE-LogE \label{sec:Comparing-Tensors}}

Therefore, from expressions (\ref{eq:LogE00})-(\ref{eq:LogE01}),
given matrices $A_{1}$ and $A_{2}$ in the form of expression (\ref{eq:matrixA}), we can compute the difference:

\[
D=\log\left(A_{1}\right)-\log\left(A_{2}\right)=
\]

\begin{equation}
\left(\begin{array}{ccc}
\log\left(L_{1}^{-T}\right) &  & \left(L_{1}^{-T}-I\right)^{-1}\log\left(L_{1}^{-T}\right)\mathbf{p}\\
\\
0^{T} &  & 0
\end{array}\right)-\left(\begin{array}{ccc}
\log\left(L_{2}^{-T}\right) &  & \left(L_{2}^{-T}-I\right)^{-1}\log\left(L_{2}^{-T}\right)\mathbf{q}\\
\\
0^{T} &  & 0
\end{array}\right).\label{eq:diff-tensors-LogE}
\end{equation}

So the matrix $D$ encompasses two kinds of non-null elements; the
first one that holds the Cholesky decomposition of the tensors:

\begin{equation}
D\left(1,1\right)=\log\left(L_{1}^{-T}\right)-\log\left(L_{2}^{-T}\right),\label{eq:D11}
\end{equation}
and the second one where the coordinates of the points $\mathbf{p}$
and $\mathbf{q}$ explicitly appear:

\begin{equation}
D\left(1,2\right)=\left(L_{1}^{-T}-I\right)^{-1}\log\left(L_{1}^{-T}\right)\mathbf{p}-\left(L_{2}^{-T}-I\right)^{-1}\log\left(L_{2}^{-T}\right)\mathbf{q}.\label{eq:D12}
\end{equation}

We follow the ICP-CTSF philosophy and assume that the points corresponding
to the same region in two different point clouds have similar
shape tensors because their surrounding geometry are
the same. Hence $D\left(1,1\right)$ in expression (\ref{eq:D11})
should be low. However, we shall include the distance between points in order to avoid convergence to local minimum. Expression (\ref{eq:D12}) can be used to solve this problem once it includes the positions of the points $\mathbf{p}$ and $\mathbf{q}$.
Therefore, we can compare the geometries nearby points $\mathbf{p}$
and $\mathbf{q}$ using the square of Frobenius norm in LogE space:
\[
\left\Vert D(\mathbf{p},\mathbf{q}) \right\Vert _{F}^{2}=
\]

\begin{equation}
\left\Vert \log\left(L_{1}^{-T}\right)-\log\left(L_{2}^{-T}\right)\right\Vert _{F}^{2}+\left\Vert \left(L_{1}^{-T}-I\right)^{-1}\log\left(L_{1}^{-T}\right)\mathbf{p}-\left(L_{2}^{-T}-I\right)^{-1}\log\left(L_{2}^{-T}\right)\mathbf{q}\right\Vert _{2}^{2},\label{eq:FrobNorm}
\end{equation}
or a variant which includes a weight $\omega$ to distinguish the
influence of each term:

\[
\left\Vert D(\mathbf{p},\mathbf{q},\omega) \right\Vert _{F\omega}^{2}=
\]

\begin{equation}
\omega \left\Vert \log\left(L_{1}^{-T}\right)-\log\left(L_{2}^{-T}\right)\right\Vert _{F}^{2} +  \left\Vert \left(L_{1}^{-T}-I\right)^{-1}\log\left(L_{1}^{-T}\right)\mathbf{p}-\left(L_{2}^{-T}-I\right)^{-1}\log\left(L_{2}^{-T}\right)\mathbf{q}\right\Vert _{2}^{2}.\label{eq:VarFrobNorm}
\end{equation}

However, before going ahead we must analyse the influence of rotations and translations in the above expressions.

\section{Effect of Rigid Transformations \label{sec:effect-rigidT}}

We consider the rigid transformation:

\begin{equation}
\rho\left(\mathbf{x}\right)=R\mathbf{x}+\mathbf{t},\label{eq:rigidTransf},
\end{equation}
and the perfect alignment where $Q=\rho(P)$. So, we can compute
the second-order tensor field defined by expression (\ref{eq:stage1}) in the target cloud $Q$:
\begin{equation}
\mathbf{T}\left(\rho\left(\mathbf{p}\right)\right)=\sum\limits _{\rho\left(\mathbf{s}\right)\in L_{k}(\rho\left(\mathbf{p}\right))}\exp\left[\dfrac{-||\mathbf{v}_{\rho\left(p\right)\rho\left(s\right)}||_{2}^{2}}{\sigma^{2}\left(\rho\left(\mathbf{p}\right)\right)}\right]\cdot\left(\widehat{\mathbf{v}}_{\rho\left(p\right)\rho\left(s\right)}\cdot\widehat{\mathbf{v}}_{\rho\left(p\right)\rho\left(s\right)}^{T}\right).\label{eq:stage1-1}
\end{equation}

Hence, given $\mathbf{p}\in P$ and $L_{k}(\mathbf{p})\subset P$, we
also have: $\mathbf{\rho\left(p\right)}\in Q$ and $L_{k}(\mathbf{\rho\left(p\right)})\subset Q$
and $\rho\left(\mathbf{s}\right)\in L_{k}(\rho\left(\mathbf{p}\right))$.
Like in section \ref{sec:icp-ctsf}, we define $\mathbf{v}_{\rho\left(p\right)\rho\left(s\right)}=\left(\rho\left(\mathbf{s}\right)-\rho\left(\mathbf{p}\right)\right)$, compute its normalized version:

\begin{equation}
\widehat{\mathbf{v}}_{\rho\left(p\right)\rho\left(s\right)}=\frac{\mathbf{v}_{\rho\left(p\right)\rho\left(s\right)}}{||\mathbf{v}_{\rho\left(p\right)\rho\left(s\right)}||_{2}}=\frac{\left(R\mathbf{s}+\mathbf{t}\right)-\left(R\mathbf{p}+\mathbf{t}\right)}{||\left(R\mathbf{s}+\mathbf{t}\right)-\left(R\mathbf{p}+\mathbf{t}\right)||_{2}}=\frac{R\left(\mathbf{s}-\mathbf{p}\right)}{||R\left(\mathbf{s}-\mathbf{p}\right)||_{2}}=\frac{R\left(\mathbf{s}-\mathbf{p}\right)}{||\mathbf{s}-\mathbf{p}||_{2}},\label{eq:Rsp}
\end{equation}
and the function:
\begin{equation}
\sigma\left(\rho\left(\mathbf{p}\right)\right)=\sqrt{\frac{||\rho\left(\mathbf{s}_{f}\right)-\rho\left(\mathbf{p}\right)||_{2}^{2}}{\ln0.01}}=\sqrt{\frac{||R\left(\mathbf{s}_{f}-\mathbf{p}\right)||_{2}^{2}}{\ln0.01}}=\sqrt{\frac{||\mathbf{s}_{f}-\mathbf{p}||_{2}^{2}}{\ln0.01}}=\sigma\left(\mathbf{p}\right),\label{eq:sigma-new}
\end{equation}
once $R$ is an orthogonal matrix.
However, from expressions (\ref{eq:rigidTransf}) and (\ref{eq:sigma-new})
we obtain:

\begin{equation}
\exp\left[\dfrac{-||\mathbf{v}_{\rho\left(p\right)\rho\left(s\right)}||_{2}^{2}}{\sigma^{2}\left(\rho\left(\mathbf{p}\right)\right)}\right]=\exp\left[\dfrac{-||R\left(\mathbf{s}-\mathbf{p}\right)||_{2}^{2}}{\sigma^{2}\left(\mathbf{p}\right)}\right]=\exp\left[\dfrac{-||\mathbf{s}-\mathbf{p}||_{2}^{2}}{\sigma^{2}\left(\mathbf{p}\right)}\right].\label{eq:rigidexp}
\end{equation}

Moreover:

\[
\widehat{\mathbf{v}}_{\rho\left(p\right)\rho\left(s\right)}\cdot\widehat{\mathbf{v}}_{\rho\left(p\right)\rho\left(s\right)}^{T}=\frac{R\left(\mathbf{s}-\mathbf{p}\right)}{||\mathbf{s}-\mathbf{p}||_{2}}\cdot\left(\frac{R\left(\mathbf{s}-\mathbf{p}\right)}{||\mathbf{s}-\mathbf{p}||_{2}}\right)^{T}=R\left[\frac{\left(\mathbf{s}-\mathbf{p}\right)\left(\mathbf{s}-\mathbf{p}\right)^{T}}{||\mathbf{s}-\mathbf{p}||_{2}^{2}}\right]R^{T},
\]
consequently:

\begin{equation}
\widehat{\mathbf{v}}_{\rho\left(p\right)\rho\left(s\right)}\cdot\widehat{\mathbf{v}}_{\rho\left(p\right)\rho\left(s\right)}^{T}=R\widehat{\mathbf{v}}_{ps}\cdot\widehat{\mathbf{v}}_{ps}R^{T}.\label{eq:RvpsR}
\end{equation}

Hence, if we insert results (\ref{eq:rigidexp}) and (\ref{eq:RvpsR})
into expression (\ref{eq:stage1-1}) we get:

\[
\mathbf{T}\left(\rho\left(\mathbf{p}\right)\right)=\sum\limits _{s\in L_{k}(\mathbf{p})}\exp\left[\dfrac{-||\mathbf{s}-\mathbf{p}||_{2}^{2}}{\sigma^{2}\left(\mathbf{p}\right)}\right]\cdot\left(R\widehat{\mathbf{v}}_{ps}\cdot\widehat{\mathbf{v}}_{ps}R^{T}\right)=R\mathbf{T}\left(\mathbf{p}\right)R^{T}.
\]

Regarding the Cholesky decomposition , we can write:

\begin{equation}
\left(R\Sigma R^{T}\right)^{-1}=R\Sigma^{-1}R^{T}=\widetilde{L}\widetilde{L}^{T}.\label{eq:Ltil}
\end{equation}
where $\widetilde{L}$ is a lower triangular matrix.

Therefore, under rigid transformation $\mathcal{N}\left(\mathbf{p},\Sigma\right)\longrightarrow\mathcal{N}\left(R\left(\mathbf{p}\right),R\Sigma R^{T}\right)$
and the DE-LogE embedding process becomes:

\begin{equation}
\mathcal{N}\left(R\mathbf{p}+\mathbf{t},R\Sigma R^{T}\right)\quad\underrightarrow{\varphi^{-1}}\quad A_{R\mathbf{p}+\mathbf{t},\widetilde{L}^{-T}}\quad\underrightarrow{\log}\quad\log\left(A_{R\mathbf{p}+\mathbf{t},\widetilde{L}^{-T}}\right)=\log\left(\begin{array}{cc}
\widetilde{L}^{-T} & R\mathbf{p}+\mathbf{t}\\
0^{T} & 1
\end{array}\right).\label{eq:DE-LogE-1}
\end{equation}

So, given $\mathbf{p}\in P$ and $\rho\left(\mathbf{p}\right)\in Q$
we obtain:
\[
D=\log\left(\begin{array}{cc}
L^{-T} & \mathbf{p}\\
0^{T} & 1
\end{array}\right)-\log\left(\begin{array}{cc}
\widetilde{L}^{-T} & R\mathbf{p}+\mathbf{t}\\
0^{T} & 1
\end{array}\right)=
\]

\begin{equation}
\left(\begin{array}{ccc}
\log\left(L^{-T}\right) &  & \left(L^{-T}-I\right)^{-1}\log\left(L^{-T}\right)\mathbf{p}\\
\\
0^{T} &  & 0
\end{array}\right)-\left(\begin{array}{ccc}
\log\left(\widetilde{L}^{-T}\right) &  & \left(\widetilde{L}^{-T}-I\right)^{-1}\log\left(\widetilde{L}^{-T}\right)\left(R\mathbf{p}+\mathbf{t}\right)\\
\\
0^{T} &  & 0
\end{array}\right).\label{eq:diff-tensors-LogE-1}
\end{equation}

In this case, the ideal situation would be $D\left(1,1\right)=0$, where $D\left(1,1\right)=\log\left(L^{-T}\right)-\log\left(\widetilde{L}^{-T}\right)$. However, we can not guarantee this property because $L \neq \widetilde{L}$  in general. As a consequence, we can not guarantee invariance to rotations for $D\left(1,1\right)$ which is a problem for registration algorithms.

%Expression (\ref{eq:diff-tensors-LogE-1}), shows that we can not guarantee $D\left(1,1\right)=\log\left(L^{-T}\right)-%\log\left(\widetilde{L}^{-T}\right)$ is invariant agains rigid transformations because $L??\widetilde{L}$  in general (??%Confirmar??).

\section{Experimental Results \label{sec:Experimental-Results}}

Despite of the difficulties regarding the non-invariance of $D\left(1,1\right)$ against  rotations, we present in this section some tests to evaluate expressions (\ref{eq:FrobNorm})-(\ref{eq:VarFrobNorm}) in rigid transformation tasks. Specifically, we keep the strategy of expression (\ref{eq:dcm}) and put together the Euclidean distances and shape features by applying expressions (\ref{eq:FrobNorm}) and (\ref{eq:VarFrobNorm}) instead of (\ref{eq:dcm}), generating two strategies:
\begin{enumerate}[start=0]
  \item  Replace expression (\ref{eq:dcm}) to equation (\ref{eq:FrobNorm}),
  \item  Replace expression (\ref{eq:dcm})  to equation (\ref{eq:VarFrobNorm}) with $\omega$ defined likewise $w_{m}$ in Algorithms \ref{alg:ICP-CTSF}-\ref{alg:SWC-ICP},
%  \item Replace expression (\ref{eq:dcm})  to equation (\ref{eq:VarFrobNorm}) with $\omega_{1}=1$ and $\omega_{2}=1/w_{m}$ where $w_{m}$ is defined likewise in Algorithm \ref{alg:ICP-CTSF}.
\end{enumerate}

%an expression analogous to equation (\ref{eq:dcm}), but replacing the CTSF term to equation (\ref{eq:FrobNorm}), that means:
%\begin{equation}
%d_{L,m}(\mathbf{p},\mathbf{q},m)= \gamma ||\mathbf{p}-\mathbf{q}||_{2}+w_{m}\cdot \left\Vert D(\mathbf{p},\mathbf{q}) \right\Vert _{F}^{2}, \label{eq:dLiem}
%\end{equation}
%where $\left\Vert D(\mathbf{p},\mathbf{q}) \right\Vert _{F}^{2}$ is given by equation (\ref{eq:FrobNorm}), $w_{m}=w_{0}b^{m}$, with $b<1$, and $0<w_{m}<w_{0}$, and $ \gamma \in \left\{ 0,1 \right\}$.

\begin{figure}[!t]
\begin{minipage}[t]{1\linewidth}%
\centering %
%\mbox{%
\includegraphics[width=0.35\linewidth]{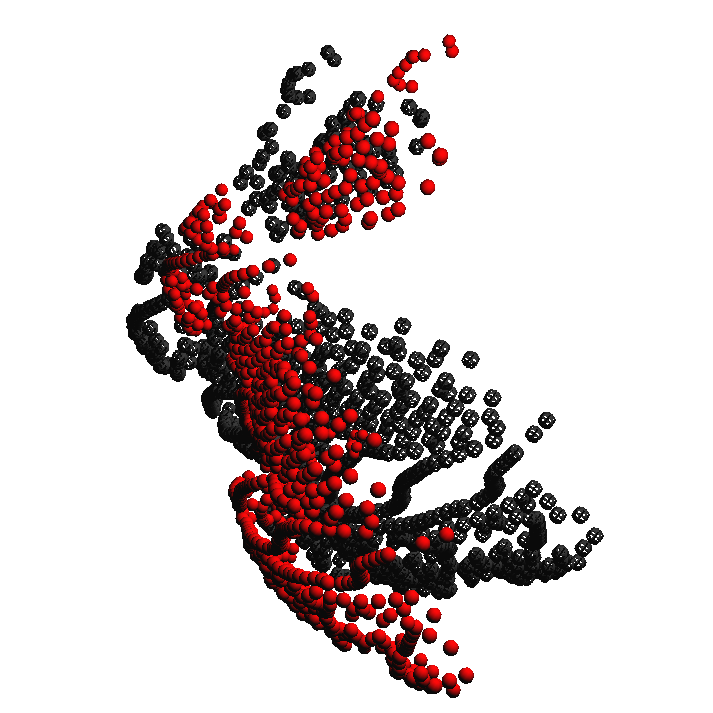}(i)%
\includegraphics[width=0.35\linewidth]{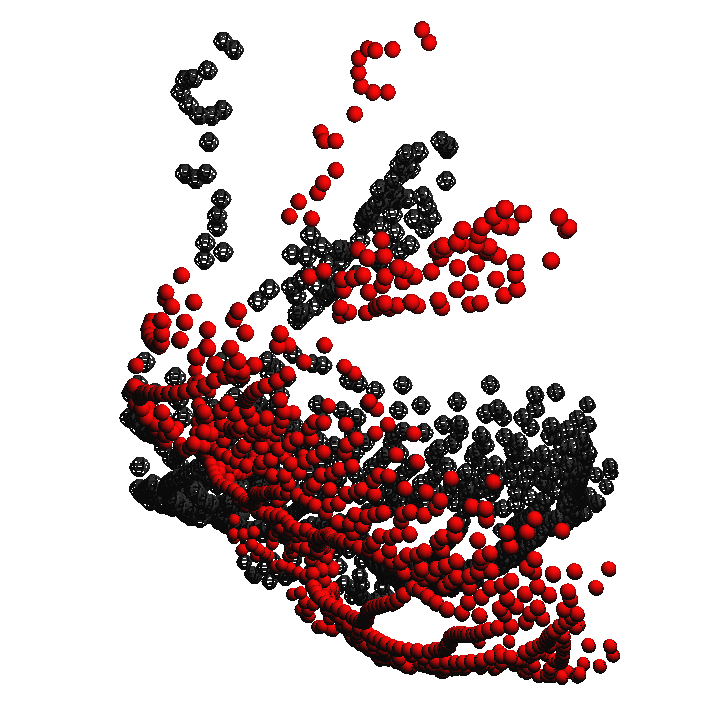}(ii)%
\par
\includegraphics[width=0.35\linewidth]{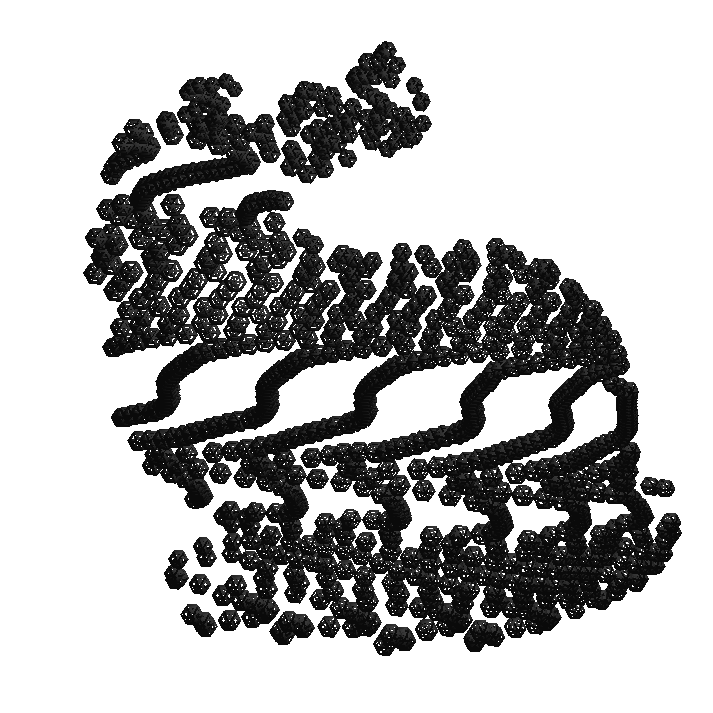}(iii)
\includegraphics[width=0.35\linewidth]{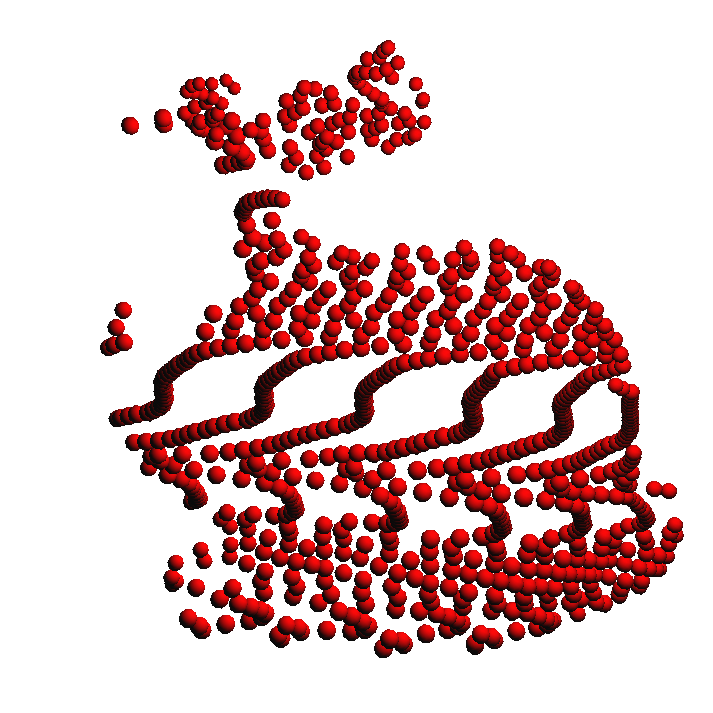}(iv)
\par
\includegraphics[width=0.35\linewidth]{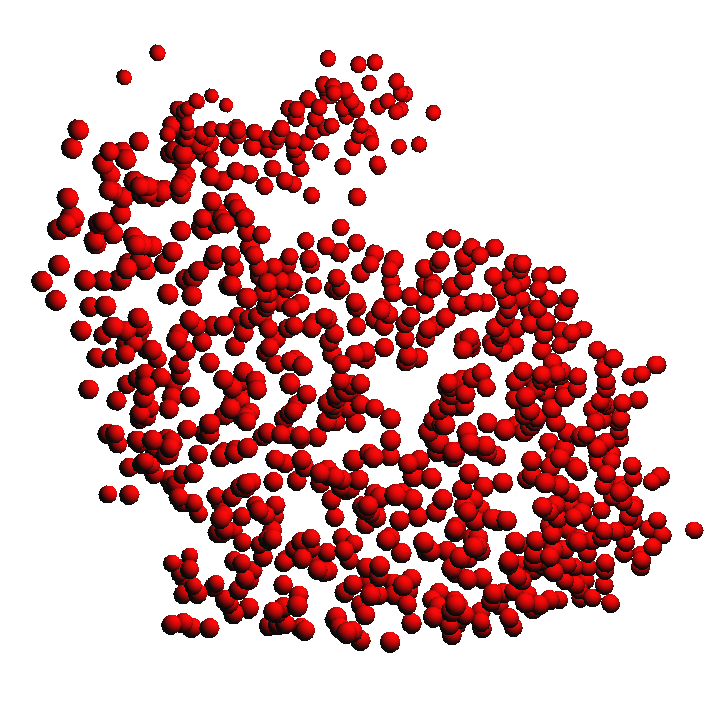}(v)
\includegraphics[width=0.35\linewidth]{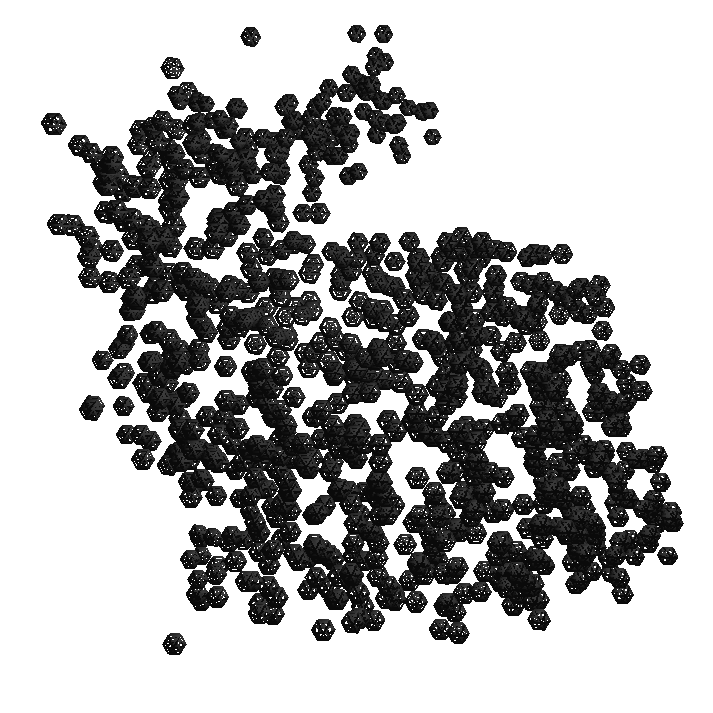}(vi)
%} %
\end{minipage}\caption{(i)-(ii) Input point clouds visualization in two point of view. The source point cloud contains red points, and the target one contains black points. (iii)-(iv) Simulating missing data (Original data and point cloud with hole). (v)-(vi) Add noise to the source and target point clouds.}
\label{fig:bunnyview}
\end{figure}

Hence, we compare the alignment given by the methods ICP-CTSF  and SWC-ICP, and  their variants using the Lie algebra, named ICP-LIE-0 and SWC-LIE-0 (case 0 above),  ICP-LIE-1 and SWC-LIE-1 (case 1 above).
The two input point clouds are obtained by sampling the Bunny model, from the Stanford
3D Scanning Repository \cite{Stanford}, based on the selection of key points at a fixed step.
For the following experiments, this step is chosen to be $45$, resulting in a set of $894$ points.
The source point cloud corresponds to the target point cloud, but rotated by $45\degree$ counterclockwise over the vertical $Y$ axis, and no translation is added (Figures \ref{fig:bunnyview}.(i)-(ii) ).
With this setup, we have the ground truth transformation and point-by-point correspondence. The registration algorithms are compared by using  root mean squared error ($MRMS$), computed by:
\begin{equation}
MRMS\left(R,\mathbf{t}\right) = \sqrt{e^{2}\left(R,\mathbf{t}\right)}, \label{eq:mrms}
%e^{2}\left(R,\mathbf{t}\right)=\frac{1}{n}\sum_{i=1}^{n}\left\Vert \mathbf{y}_{i}-\left(R\mathbf{x}_{i}+\mathbf{t}\right)\right\Vert _{2}^{2},\label{eq:eq2}
\end{equation}
where $e^{2}$ is given by equation (\ref{eq:eq2}) and the pair $\left(R,\mathbf{t}\right)$ is the rotation and translation (respectively) obtained through the registration process.

Besides the point clouds in Figures \ref{fig:bunnyview}.(i)-(ii) , we take the $0^\circ$ Bunny model configuration and build a case-study for registration under missing data by moving it using the same rotation and translation as before.
Then, we remove a set of points inside a ball centered in a specific point in the cloud, with  $radius = 0.03$, and update the correspondence set (see Figures \ref{fig:bunnyview}.(iii)-(iv)). The last scenario is composed by the original clouds corrupted by noise. Noise is simulated adding to each point a vector $\mathbf{r} = \nu \cdot \vartheta \cdot \mathbf{u}$, where $\mathbf{u}$ is a random normalized isotropic vector, $\vartheta$ is a Gaussian random variable with null mean and variance equals to $1.0$, and $\nu$ denotes a scale factor. Specifically, if $\mathbf{x}$ denotes a generic point in the set $P$ (or $Q$) then its corrupted version is $\mathbf{\widehat{x}}$, given by: $\mathbf{\widehat{x}} = \mathbf{x} + \mathbf{r}$.
The new point sets, denoted by $P_{\nu}$ and $Q_{\nu}$, are composed by the noisy points so generated. We experiment with $\nu = 5\%$ to generate the noisy clouds $P_{\nu}$ and $Q_{\nu}$, which are pictured on Figures \ref{fig:bunnyview}.(v)-(vi).

\begin{figure}[!t]
\begin{minipage}[t]{1\linewidth}%
\centering %
%\mbox{%
\includegraphics[width=0.45\linewidth]{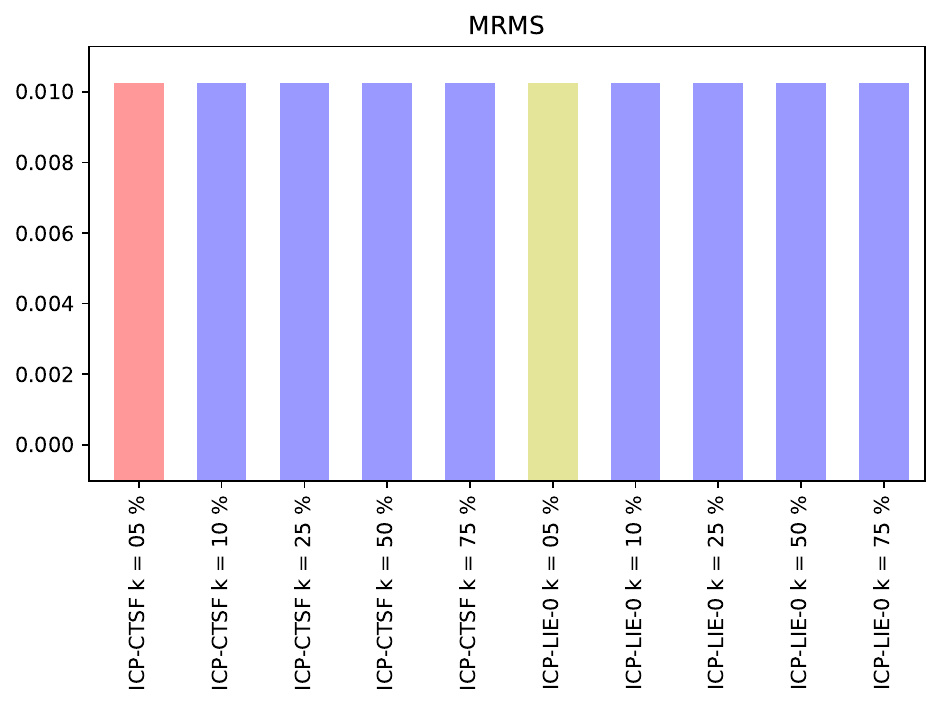}(a)
\includegraphics[width=0.45\linewidth]{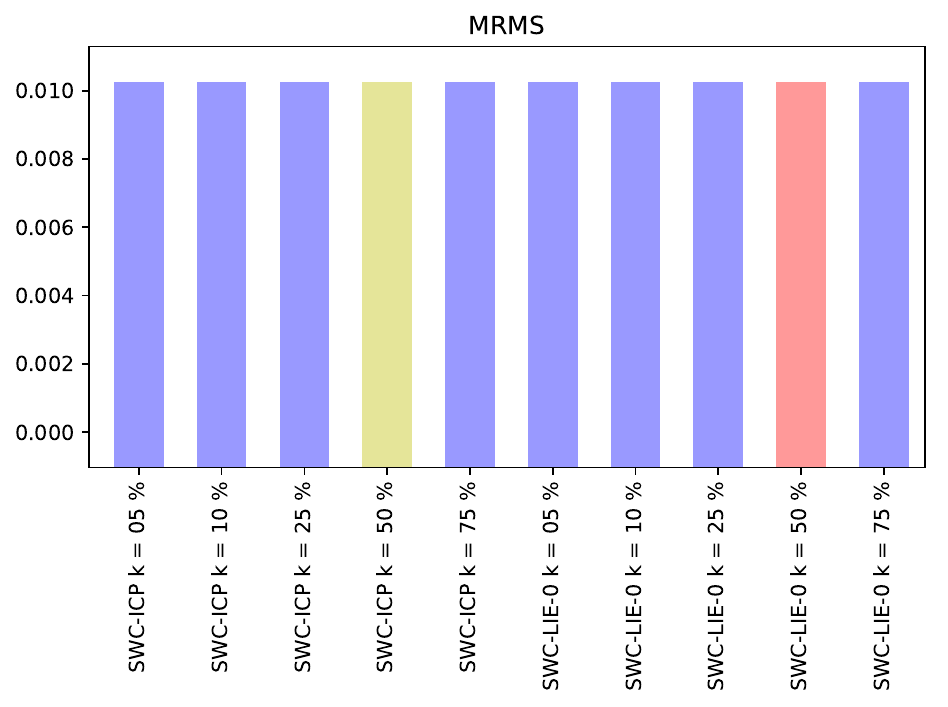}(b)
%}
\par
\includegraphics[width=0.45\linewidth]{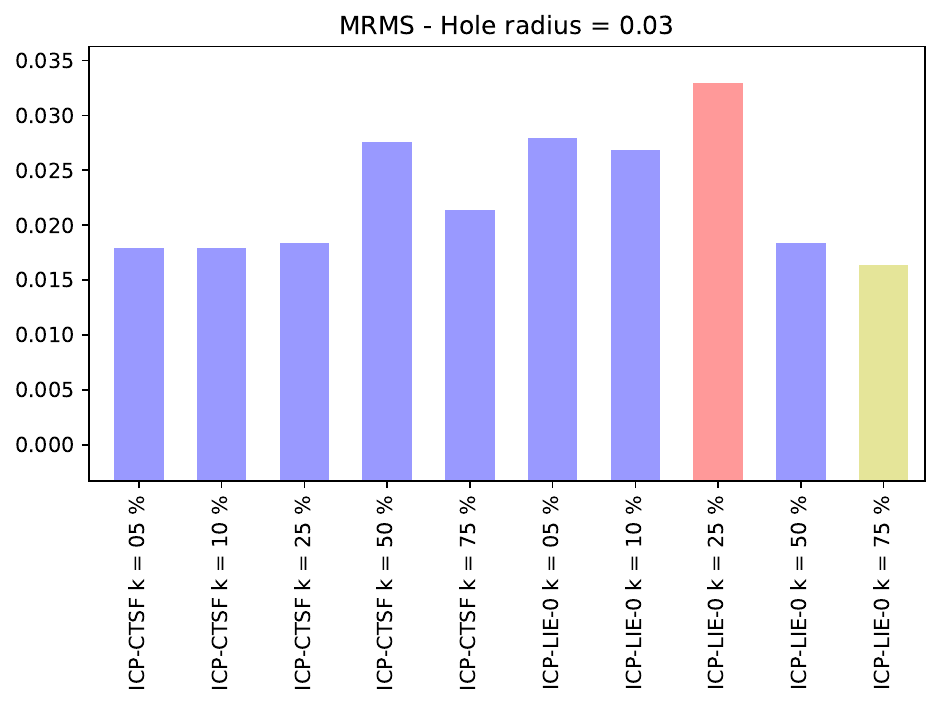}(c)
\includegraphics[width=0.45\linewidth]{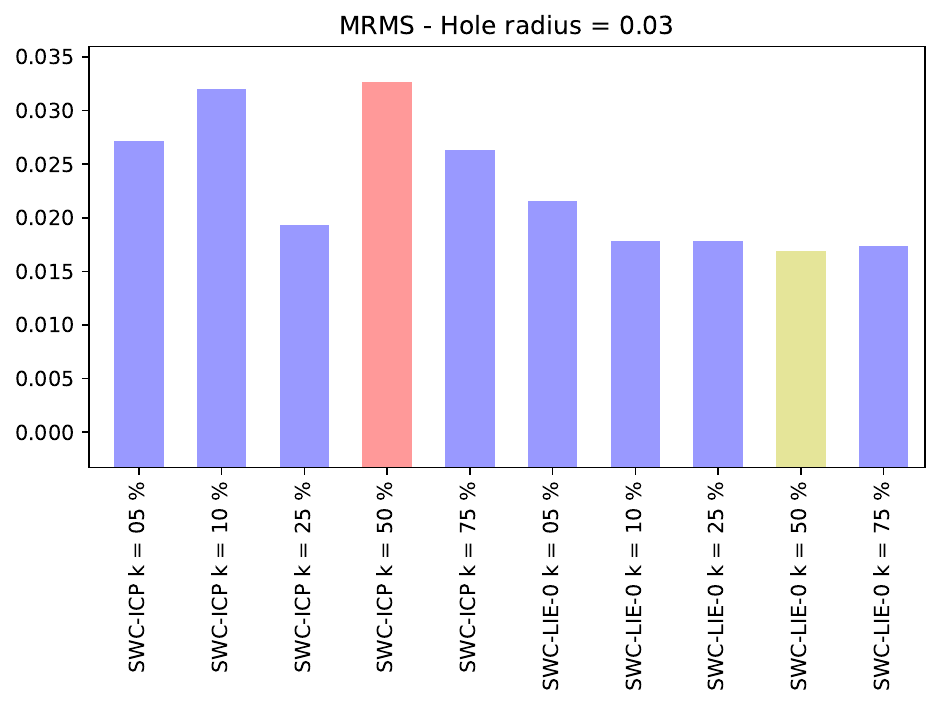}(d)
\par
\includegraphics[width=0.45\linewidth]{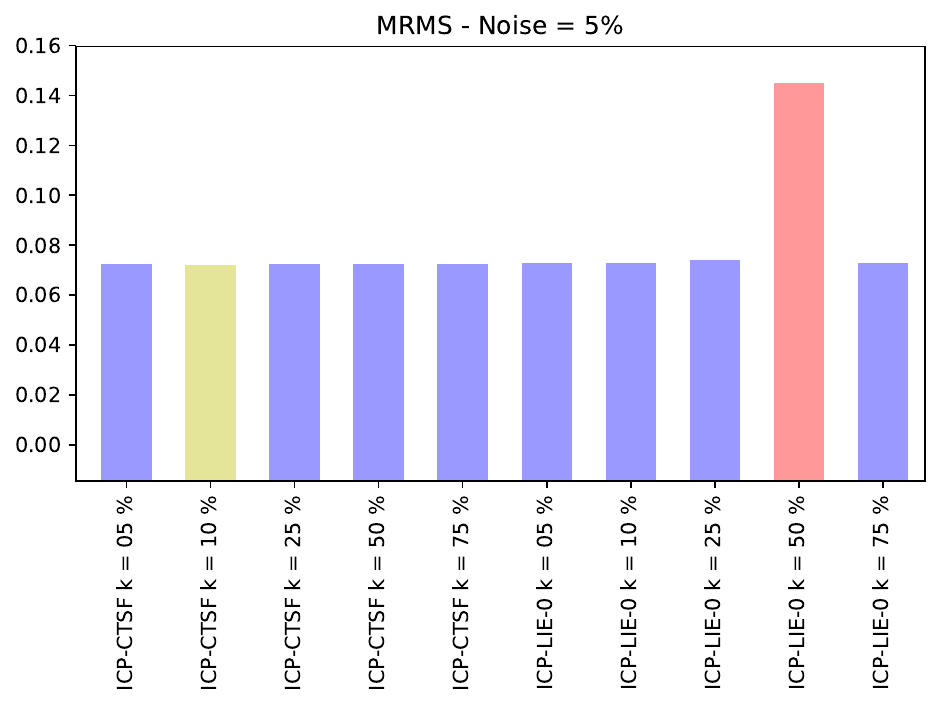}(e)
\includegraphics[width=0.45\linewidth]{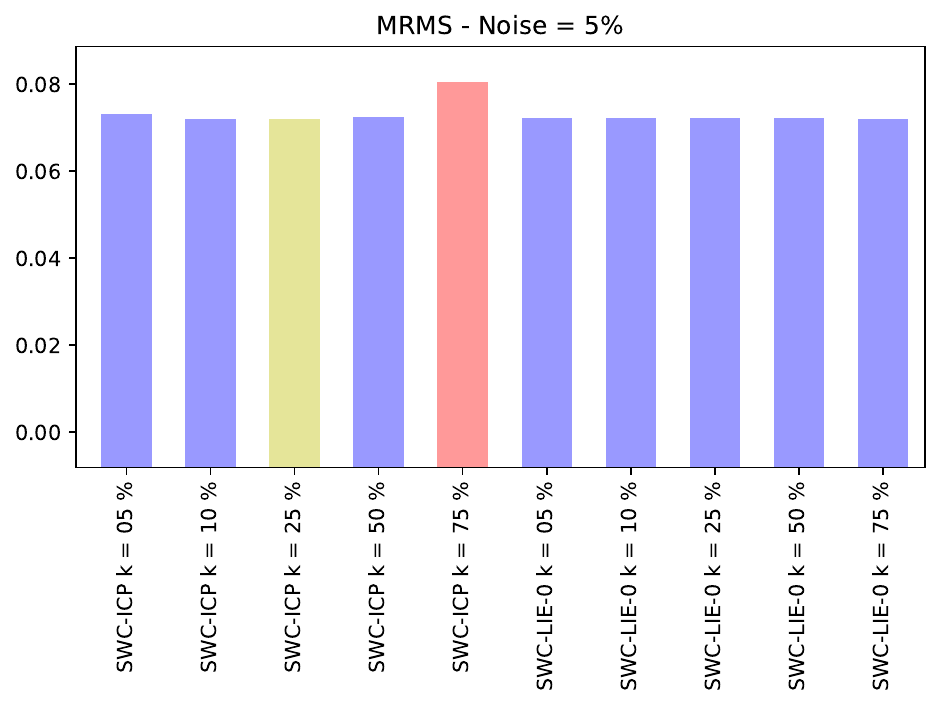}(f)
\end{minipage}
\caption{$MRMS$ of registration algorithms using expression (\ref{eq:FrobNorm}). The minimum value is highlighted in yellow, and the maximum value, in red. Source-target clouds are given in Figure \ref{fig:bunnyview}.(i)-(ii): (a) ICP-CTSF versus ICP-LIE-0. (b) SWC-ICP versus SWC-LIE-0. Source-target clouds given by Figures \ref{fig:bunnyview}.(iii)-(iv): (c) ICP-CTSF versus ICP-LIE-0. (d) SWC-ICP versus SWC-LIE-0. Source-target clouds defined by noisy clouds in Figures \ref{fig:bunnyview}.(v)-(vi): (e) ICP-CTSF versus ICP-LIE-0. (f) SWC-ICP versus SWC-LIE-0.}
\label{fig:ICPwithEuclidean}
\end{figure}

Figures \ref{fig:ICPwithEuclidean}.(a)-(f) allow to compare the focused techniques when applied to register clouds in Figure \ref{fig:bunnyview}. These plots report the $MRMS$ of ICP-CTSF and SWC-ICP procedures (Algorithms \ref{alg:ICP-CTSF}-\ref{alg:SWC-ICP}) according to the ground truth correspondence, as well as the $MRMS$ of each technique obtained when using strategies $0$ above.

\begin{figure}[!t]
\begin{minipage}[t]{1\linewidth}%
\centering %
%\mbox{%
\includegraphics[width=0.45\linewidth]{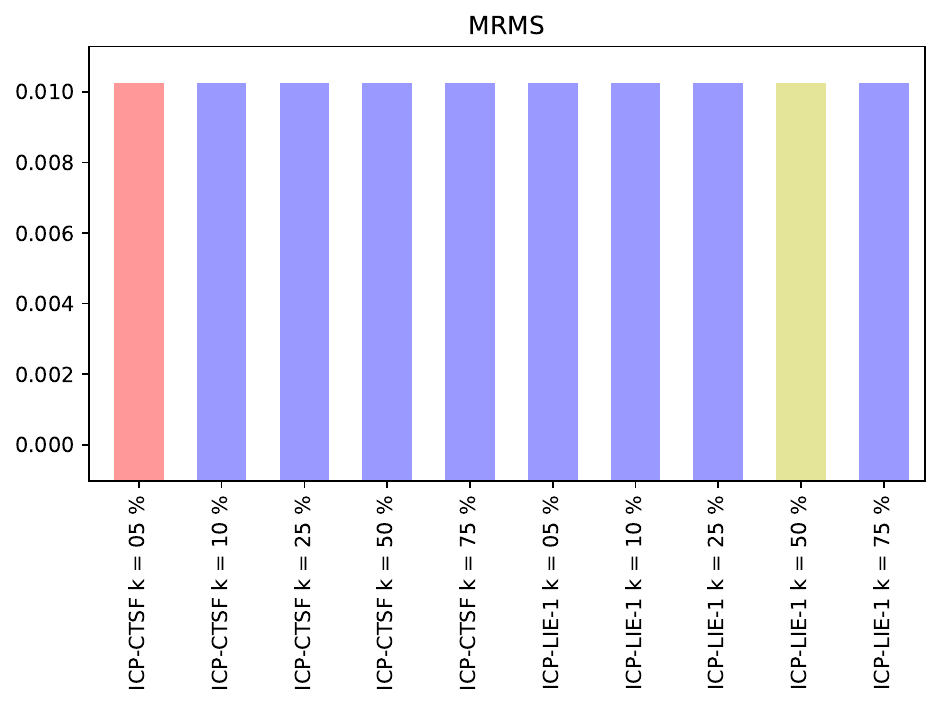}(a)
\includegraphics[width=0.45\linewidth]{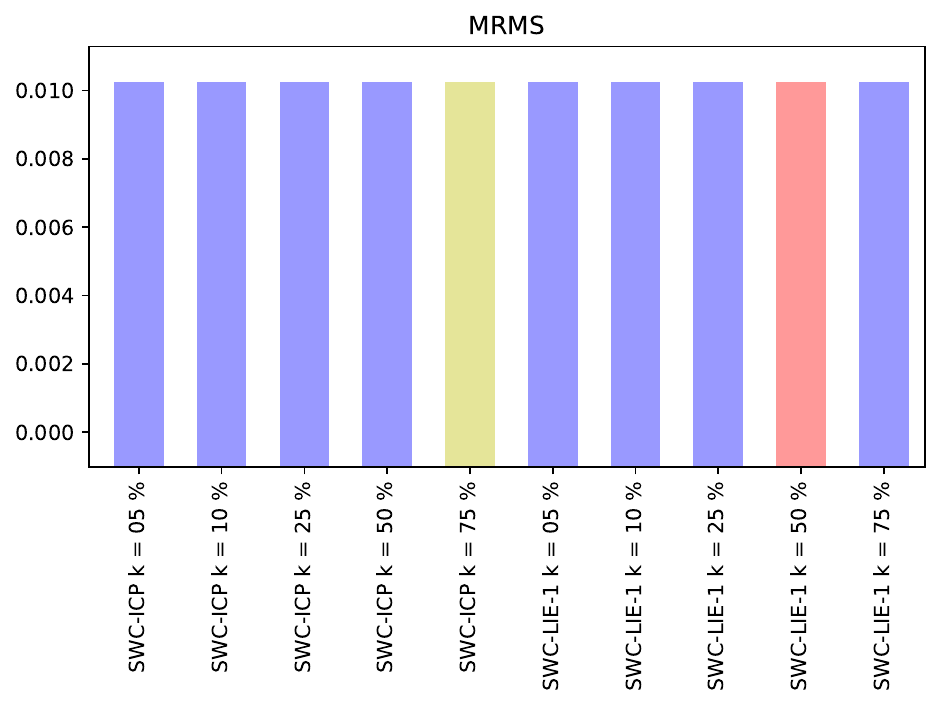}(b)
\par
\includegraphics[width=0.45\linewidth]{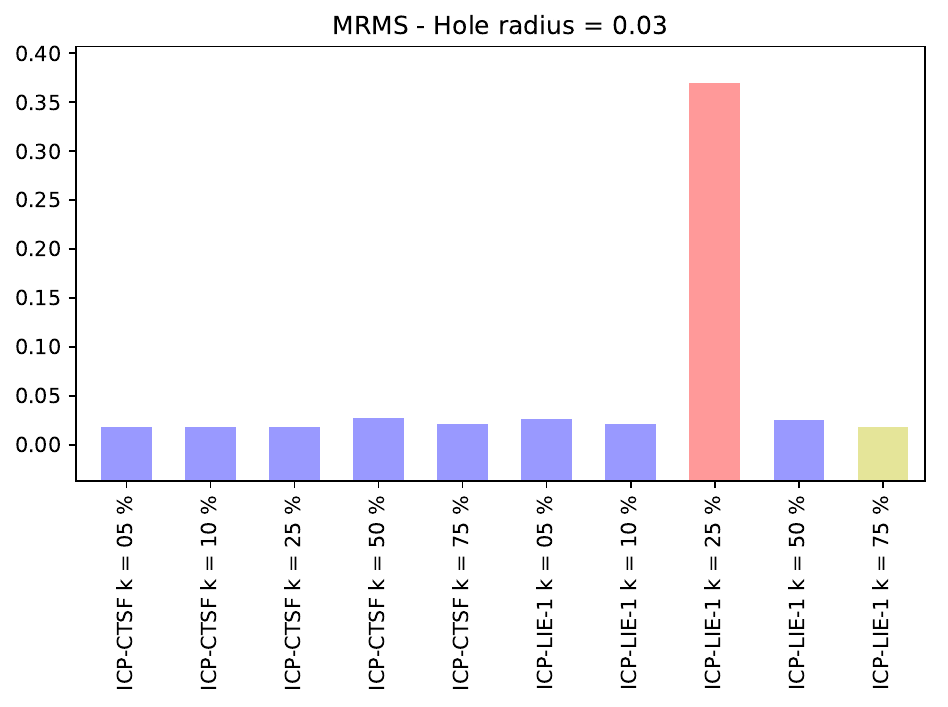}(c)
\includegraphics[width=0.45\linewidth]{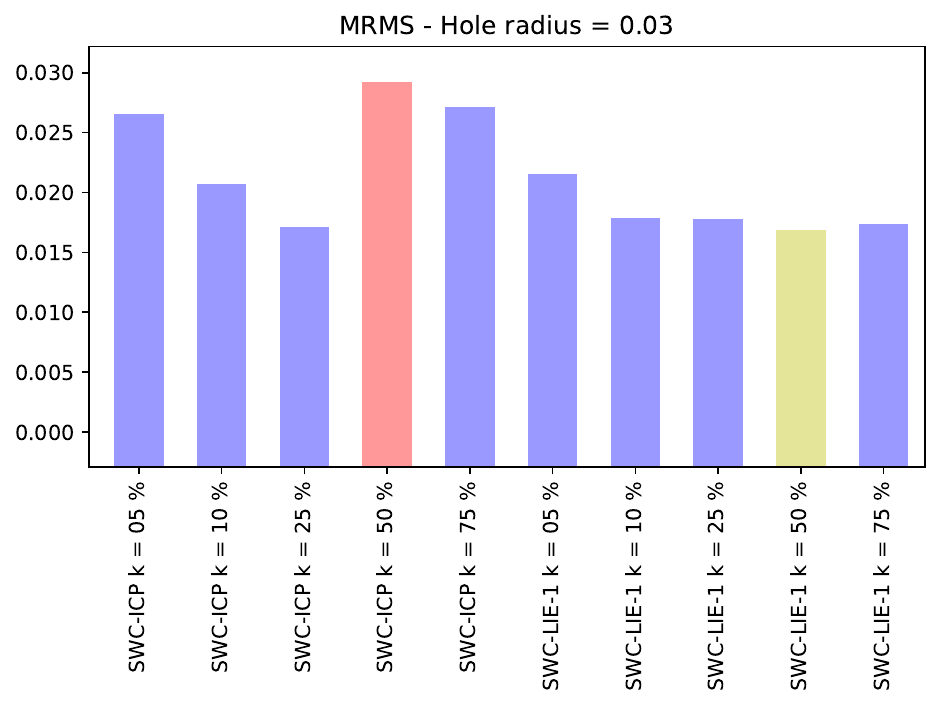}(d)
\par
\includegraphics[width=0.45\linewidth]{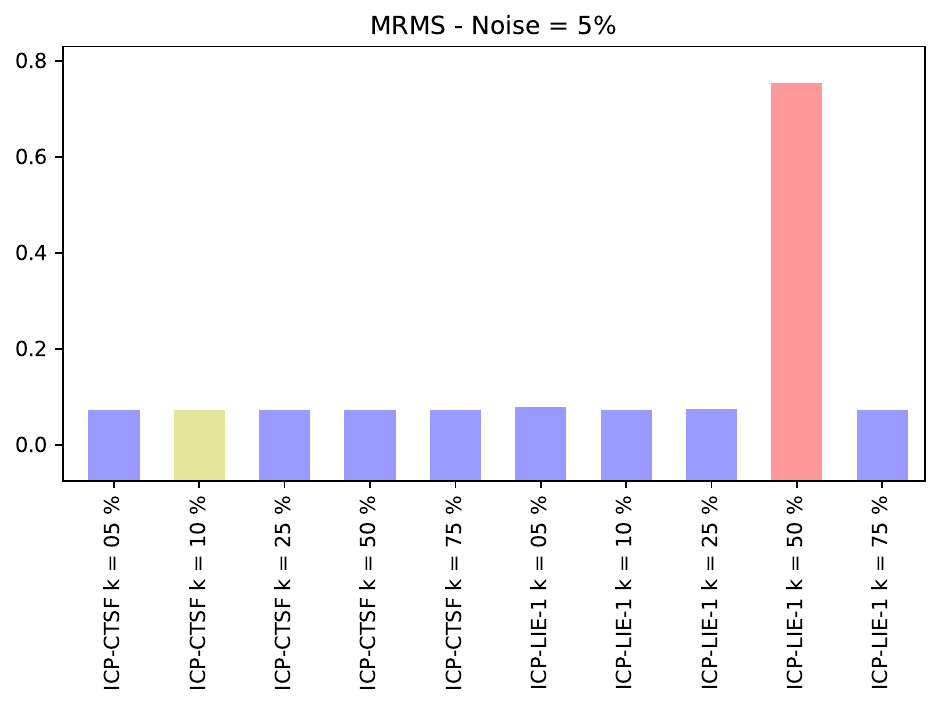}(e)
\includegraphics[width=0.45\linewidth]{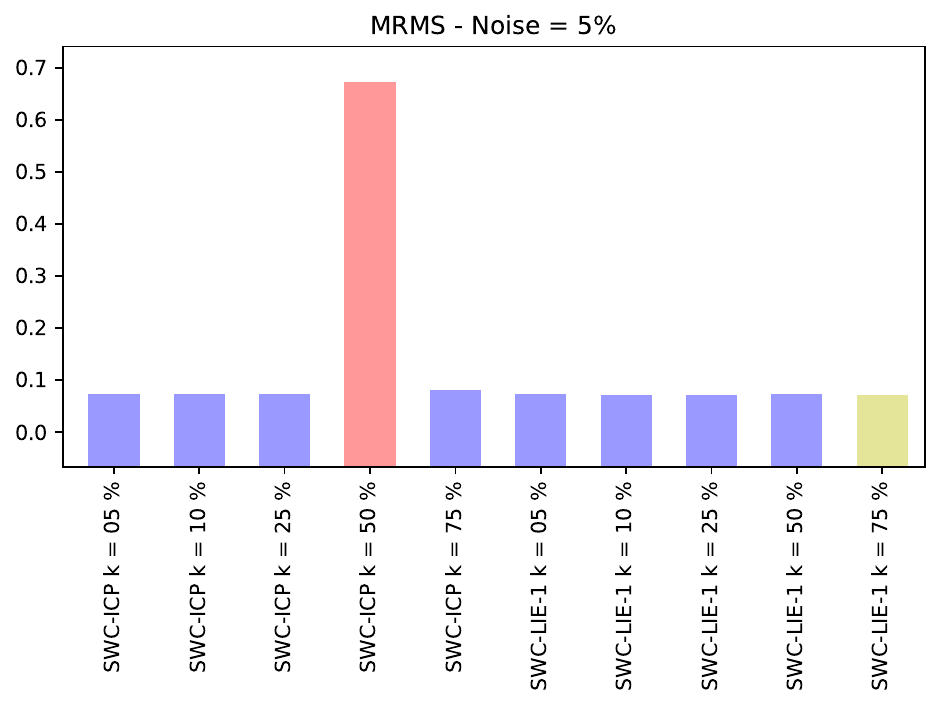}(f)
%}
\end{minipage}
\caption{$MRMS$ of registration algorithms using expression (\ref{eq:VarFrobNorm}). The minimum value is highlighted in yellow, and the maximum value, in red. Source-target clouds given in Figures \ref{fig:bunnyview}.(i)-(ii): (a) ICP-CTSF versus ICP-LIE-1. (b) SWC-ICP versus SWC-LIE-1. Source-target clouds defined by Figures \ref{fig:bunnyview}.(iii)-(iv):  (c) ICP-CTSF versus ICP-LIE-1. (d) SWC-ICP versus SWC-LIE-1. Source-target clouds given by Figures \ref{fig:bunnyview}.(v)-(vi): (e) ICP-CTSF versus ICP-LIE-1. (f) SWC-ICP versus SWC-LIE-1.}
\label{fig:ICPnoEuclidean}
\end{figure}

Next, Figures \ref{fig:ICPnoEuclidean}.(a)-(f) present the analogous results but using ICP-CTSF and SWC-ICP, together with their Lie versions obtained by replacing expression (\ref{eq:dcm})  to equation (\ref{eq:VarFrobNorm}) with $\omega$ defined likewise $w_{m}$ in Algorithms \ref{alg:ICP-CTSF}-\ref{alg:SWC-ICP}.

The Table \ref{tab:Best-Alignments-Case0-1-2} summarizes the results shown in Figures \ref{fig:ICPwithEuclidean}-\ref{fig:ICPnoEuclidean}. We notice that the Lie approaches perform better than the traditional ICP-CTSF and SWC-ICP for original clouds and in the presence of missing data (hole). However, in the scenario containing noise the ICP-CTSF outperforms all the other approaches.
\begin{table}[!htb]
\centering
\begin{scriptsize}
\caption{Performance of ICP-CTSF, SWC-ICP and Lie versions using for strategies
$0$ and $1$ defined above. The best methods are highlighted
in bold and their $MRMS$ are reported in the last column.}
\label{tab:Best-Alignments-Case0-1-2}
\def\arraystretch{1.5}
\begin{tabular}{|c|c|c|c|c|}
\hline
Strategies  & Scenario  & Best ICP  & Best SWC  & Best $MRMS$ \tabularnewline
\hline
\hline
% ICP 5%, 10%, 50% 75% = 0.010257717275893   SWC 50% = 0.010257717297312
\multirow{3}{*}{0}  & Original  & \textbf{ICP-LIE-0}, $k=5\%$ & SWC-ICP, $k=50\%$ & 0.010257717275893 \tabularnewline \cline{2-5}
% ICP 75% = 0.016357583571102   SWC 50% = 0.016911773470017
 & Hole  & \textbf{ICP-LIE-0}, $k=75\%$ & SWC-LIE-0, $k=50\%$ & 0.016357583571102 \tabularnewline \cline{2-5}
% ICP 10% = 0.071996330253377   SWC 25% = 0.071999208093164
 & Noise $=5\%$  & \textbf{ICP-CTSF}, $k=10\%$ & SWC-ICP, $k=25\%$ & 0.071996330253377 \tabularnewline \cline{2-5}
\hline
\hline
% ICP 50% = 0.010257717275893   SWC 75% = 0.010257717761012
\multirow{3}{*}{1}  & Original  & \textbf{ICP-LIE-1}, $k=50\%$ & SWC-ICP, $k=75\%$ & 0.010257717275893 \tabularnewline \cline{2-5}
% ICP 75% = 0.017814551467704   SWC 50% = 0.016911773470017
 & Hole  & ICP-LIE-1, $k=75\%$ & \textbf{SWC-LIE-1}, $k=50\%$ & 0.016911773470017 \tabularnewline \cline{2-5}
% ICP 10% = 0.071996330253377   SWC 75% = 0.072105135757243
 & Noise $=5\%$  & \textbf{ICP-CTSF}, $k=10\%$& SWC-LIE-1, $k=75\%$ & 0.071996330253377 \tabularnewline \cline{2-5}
\hline
%\hline
% &  &  &  & \tabularnewline
%\hline
% &  &  &  & \tabularnewline
%\hline
% &  &  &  & \tabularnewline
%\hline
%\hline
% &  &  &  & \tabularnewline
\end{tabular}\end{scriptsize}
\end{table}

\section{Conclusion and Future Works \label{sec:concl}}

In this paper we consider the pairwise rigid registration of point clouds. We take two techniques, named by the acronyms
ICP-CTSF and SWC-ICP, and  apply a Lie algebra framework to compute similarity measures used in the matching step.
We use point clouds with known ground truth transformations which allows to compare the new techniques with their original implementations. The results show that the $MRMS$ of Lie approaches are better than traditional ones for original clouds and in the presence of missing data but the traditional ICP-CTSF outperforms Lie techniques for noisy data.
Further works must be undertaken to improve ICP-LIE and SWC-LIE and to bypass the non-invariance  of DE-LogE against  rotations.

% ----------------------------------------------------------
% Referências bibliográficas
% ----------------------------------------------------------
%\bibliographystyle{plain}      % mathematics and physical sciences
\bibliography{Registration-LieGroups}
% ----------------------------------------------------------
% Glossário
% ----------------------------------------------------------
%
% Há diversas soluções prontas para glossário em LaTeX.
% Consulte o manual do abnTeX2 para obter sugestões.
%
%\glossary

\end{document}